%% file: paper.tex
\documentclass{article}

\usepackage{microtype}
\usepackage{graphicx}
\usepackage{subfigure}
\usepackage{booktabs} % for professional tables

\usepackage{hyperref}

\usepackage[accepted]{icml2021}

\input{math_commands.tex}

\usepackage{booktabs}
\usepackage[dvipsnames]{xcolor}
\usepackage{amssymb}% http://ctan.org/pkg/amssymb
\usepackage{pifont}% http://ctan.org/pkg/pifont
\usepackage{dsfont}
\usepackage{natbib}
\usepackage{appendix}
\usepackage{theorem}

\newtheorem{proposition}{Proposition}

\icmltitlerunning{Environment Inference for Invariant Learning}

\begin{document}

\twocolumn[
\icmltitle{Environment Inference for Invariant Learning}

\begin{icmlauthorlist}
\icmlauthor{Elliot Creager}{to,ve}
\icmlauthor{J\"orn-Henrik Jacobsen}{to,ve}
\icmlauthor{Richard Zemel}{to,ve}
\end{icmlauthorlist}

\icmlaffiliation{to}{University of Toronto}
\icmlaffiliation{ve}{Vector Institute}

\icmlcorrespondingauthor{Elliot Creager}{creager@cs.toronto.edu}

\icmlkeywords{Invariant Learning, Out-of-distribution Generalization, Deep Learning, Machine Learning}

\vskip 0.3in
]

\printAffiliationsAndNotice{}

\begin{abstract}
Learning models that gracefully handle distribution shifts is central to research on domain generalization, robust optimization, and fairness. A promising formulation is domain-invariant learning, which identifies the key issue of learning which features are domain-specific versus domain-invariant. An important assumption in this area is that the training examples are partitioned into ``domains'' or ``environments''. Our focus is on the more common setting where such partitions are not provided. We propose EIIL, a general framework for domain-invariant learning that incorporates Environment Inference to directly infer partitions that are maximally informative for downstream Invariant Learning. We show that EIIL outperforms invariant learning methods on the CMNIST benchmark without using environment labels, and significantly outperforms ERM on worst-group performance in the Waterbirds and CivilComments datasets. Finally, we establish connections between EIIL and algorithmic fairness, which enables EIIL to improve accuracy and calibration in a fair prediction problem.
\end{abstract}

\section{Introduction}

%%%%%%%%%%%%%%%%%%%%%%%%%%%%%%%%%%%%%%%%%%%%%%%%%%%%%%%%%%%%
% fig 1
%%%%%%%%%%%%%%%%%%%%%%%%%%%%%%%%%%%%%%%%%%%%%%%%%%%%%%%%%%%%
\begin{figure}[t!]
\centering
\subfigure[\textbf{Inferred environment 1}  \textit{(mostly) landbirds on land, and waterbirds on water}]{
\includegraphics[width=.45\columnwidth]{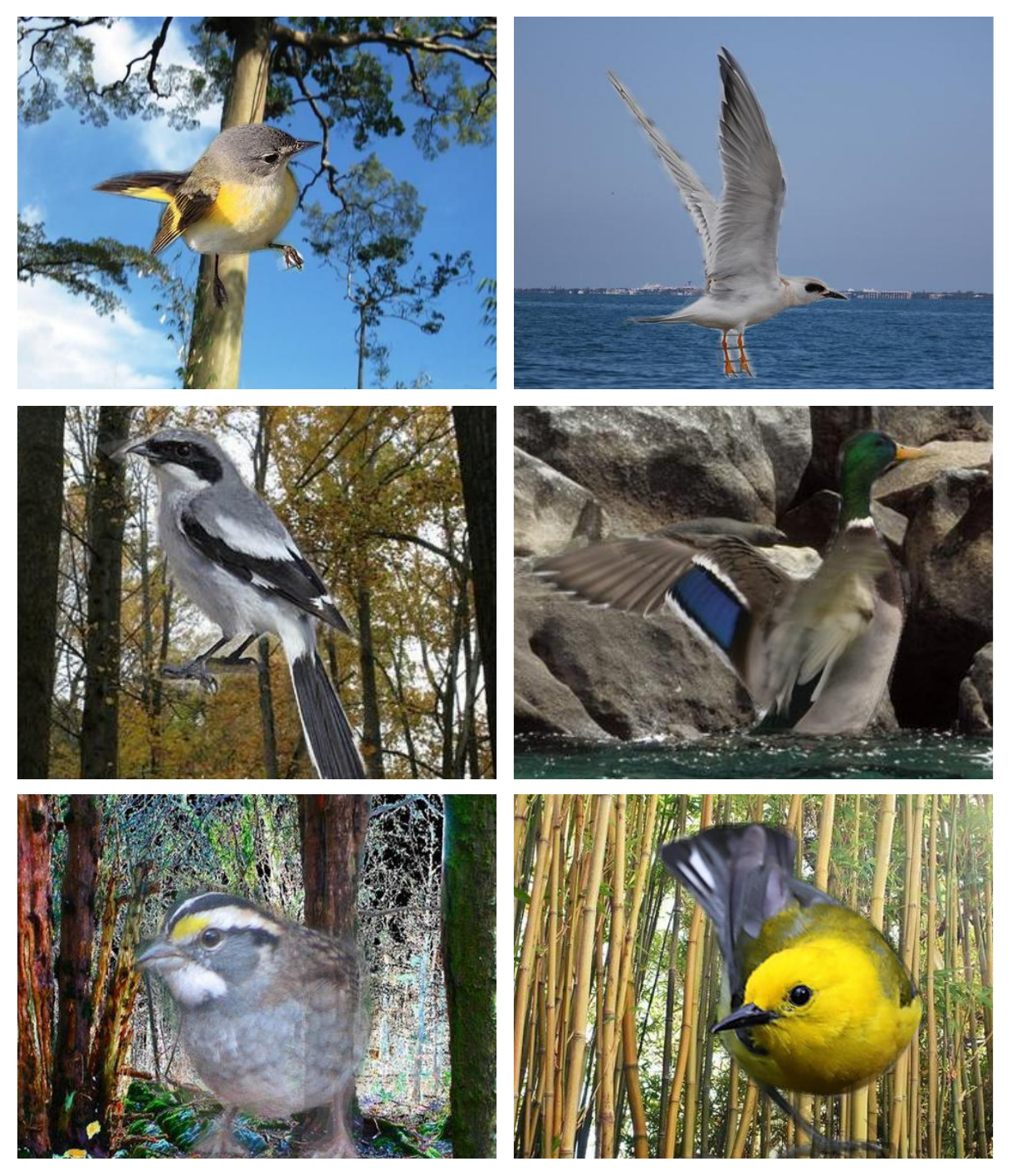}
}
\hspace{.2cm}
\subfigure[\textbf{Inferred environment 2}  \textit{(mostly) landbirds on water, and waterbirds on land}]{
\includegraphics[width=.45\columnwidth]{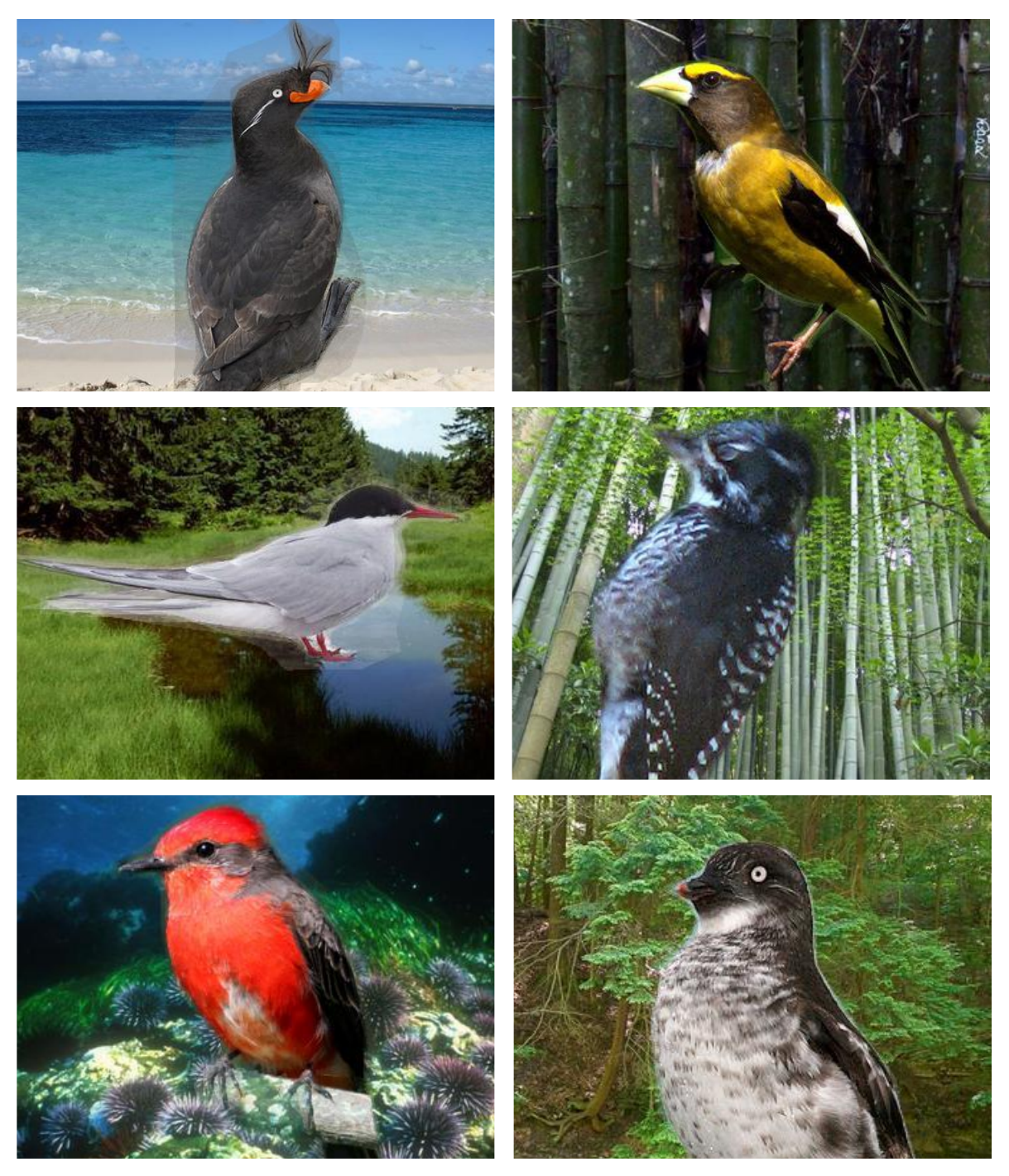}
}
\caption[]{
In the Waterbirds dataset \citep{sagawa2019distributionally}, the two target labels (landbirds and waterbirds) are correlated with
their respective typical background habitats (land and water).
This spurious correlation causes sub-par performance on the smallest subgroups (e.g. waterbirds on land). 
Environment Inference for Invariant Learning (EIIL) organizes the training data into two environments that are maximally informative for use by a downstream invariant learner, enabling the use of invariant learning in situations where environment labels are not readily available.
By grouping examples where class and background disagree into the same environment, EIIL encourages learning an invariance w.r.t. background features, which improves worst-group test accuracy by 18\% relative to standard supervised learning.
}
\label{fig:fig1}
\end{figure}
%%%%%%%%%%%%%%%%%%%%%%%%%%%%%%%%%%%%%%%%%%%%%%%%%%%%%%%%%%%%

Machine learning achieves super-human performance on many tasks when the test data is drawn from the same distribution as the training data. However, when the two distributions differ, model performance can severely degrade, even to below-chance predictions~\citep{geirhos2020shortcut}. Tiny perturbations can derail classifiers, as shown by adversarial examples~\citep{szegedy2013intriguing} and common image corruptions~\citep{hendrycks2019benchmarking}.
Even new test sets collected from the same data acquisition pipeline induce distribution shifts that significantly harm
performance~\citep{recht2019imagenet, engstrom2020identifying}. 
Many approaches have been proposed to overcome the brittleness of supervised learning---e.g.  Empirical Risk Minimization (ERM)---in the face of distribution shifts.
Robust optimization aims to achieve good performance on any distribution close to the training distribution~\citep{goodfellow2014explaining,duchi2021statistics,madry2017towards}. Invariant learning on the other hand tries to go one step further, to generalize to distributions potentially far away from the training distribution.

However, common invariant learning methods typically come at a serious disadvantage: they require datasets to be partitioned into multiple domains or environments.\footnote{We use ``domains'', ``environments'' and ``groups''/``subgroups'' interchangeably.}
Environment assignments should implicitly define variation the algorithm should become invariant or robust to, but often such environment labels are unavailable at training time, either because they are difficult to obtain or due to privacy limitations. 
In some cases, relevant side-information or metadata, e.g., human annotations, or device ID used to take a medical image, hospital or department ID, etc., may be abundant, but it remains unclear how best to specify environments based on this information \citep{srivastava2020robustness}.
A similar issue arises in mitigating algorithmic unfairness, where so-called sensitive attributes may be difficult to define in practice \citep{hanna2020towards}, or their values may be impossible to collect.
We aim to overcome the difficulty of manual environment specification by developing a new method inspired by fairness approaches for unknown group memberships~\citep{kim2019multiaccuracy,lahoti2020fairness}.

The core idea is to leverage the bias of an ERM-trained reference model to discover useful environment partitions directly from the training data.
We derive an environment inference objective that maximizes variability across environments, and is differentiable w.r.t. a distribution over environment assignments.
After performing environment inference given a fixed reference classifier, we use the inferred environments to train an invariant learner from scratch.

Our method, \textbf{Environment Inference for Invariant Learning (EIIL)}, discovers environment labels that can then be used to train any off-the-shelf invariant learning algorithm in applications where environment labels are unavailable.
This approach can outperform ERM in settings where standard learning tends to focus on spurious features or exhibit performance discrepancies between subgroups of the training data (which need not be specified ahead of time).
EIIL discovers environments capturing spurious correlations hidden in the dataset (see Figure \ref{fig:fig1}), making them readily available for invariant learning.
Surprisingly, even when manual specification of environments is available (e.g. the CMNIST benchmark), inferring environments directly from aggregated data may \emph{improve} the quality of invariant learning.

Our main contributions are as follows:
\begin{itemize}
    \item We propose a general framework for inferring environments from data based on the bias of a reference classifier;
    \item we provide a theoretical characterization of the dependence on the reference classifier, and when we can expect the method to do well;    
    \item we derive a specific instance of environment inference in this framework using gradients w.r.t. soft environment assignments, which outperforms invariant learning (using environment labels) on the CMNIST benchmark and outperforms ERM on Waterbirds;
    \item we establish a connection to similar themes in the fairness literature, and show that our method can improve accuracy and calibration in a fair prediction problem.
\end{itemize}

\section{Invariant Learning}\label{sec:exchanging-lessons}
This section discusses the problem setting and presents background materials that will be used to formulate our proposed method.
Our approach is primarily motivated by recent approaches to learning domain- or environment-invariant representations---which we simply refer to as ``invariant learning''---that have been applied to domain adaptation and generalization tasks.

\paragraph{Notation}
%%%%%%%%%%%%%%%%%%%%%%%%%%%%%%%%%%%%%%%%%%%%%%%%%%%%%%%%%%%% Notation macros
\newcommand{\pobs}{p^{obs}}
\newcommand{\phand}{p^{obs}}
\newcommand{\Xsp}{\mathcal{X}}
\newcommand{\Ysp}{\mathcal{Y}}
\newcommand{\Zsp}{\mathcal{H}}
\newcommand{\Esp}{\mathcal{E}^{obs}}
\newcommand{\RR}{\mathbb{R}}
\newcommand{\Normal}{\mathcal{N}}
\newcommand{\obs}{\mathbf{x}} % observations
\newcommand{\cf}{\mathbf{v}} % causal feature
\newcommand{\ncf}{\mathbf{z}} % non-causal feature
\newcommand{\pcf}{\hat{\mathbf{w}_v}} % predicted causal feature
\newcommand{\pncf}{\hat{\mathbf{w}_z}} % predicted non-causal feature
\newcommand{\lbl}{\mathbf{y}} % label
\newcommand{\spa}{\RR^{N}} % space of observations
\newcommand{\hsp}{\RR^{N/2}} % half-space of observations
%%%%%%%%%%%%%%%%%%%%%%%%%%%%%%%%%%%%%%%%%%%%%%%%%%%%%%%%%%%%
Let $\Xsp$ be the input space, $\Esp$ the set of training environments (a.k.a. ``domains''), $\Ysp$ the target space.
Let $x, y, e \sim \pobs(x, y, e)$ 
be observational data, with $x \in \Xsp$, $y \in \Ysp$, and $e \in \Esp$.
$\Zsp$ denotes a representation space, from which a classifier $w \circ \Phi$ (that maps to the logit space of $\Ysp$ via a linear map $w$) can be applied.
${\Phi: \Xsp \rightarrow \Zsp}$ denotes the parameterized mapping or ``model'' that we optimize.
We refer to $\Phi(x) \in \Zsp$ as the ``representation'' of example $x$.
$\hat y \in \Ysp$ denotes a hard prediction derived from the classifier by stochastic sampling or probability thresholding.
$\ell: \Zsp \times \Ysp \rightarrow \R$ denotes a scalar loss, which guides the learning.

The empirical risk minimization (ERM) solution is found by minimizing the global risk, expressed as the expected loss over the observational distribution:
\begin{equation}
    C^{ERM}(\Phi) = \E_{\pobs(x, y, e)}[\ell(\Phi(x), y)].
\end{equation}

\paragraph{Representation Learning with Environment Labels}
Domain generalization is concerned with achieving low error rates on unseen test distributions $p(x, y|e_{test})$ for $e_{test} \notin \Esp$.
Domain adaption is a related problem where model parameters can be adapted at test time using unlabeled data.
Recently, \emph{Invariant Learning} approaches such as Invariant Risk Minimization (IRM) \citep{arjovsky2019invariant} and Risk Extrapolation (REx) \citep{krueger2020out} were proposed to overcome the limitations of adversarial domain-invariant representation learning \citep{zhao2019learning} by discovering invariant relationships between inputs and targets across domains.
Invariance serves as a proxy for causality, as features representing ``causes'' of target labels rather than effects will generalize well under intervention.
In IRM, a representation $\Phi(x)$ is learned that performs optimally within each environment---and is thus invariant to the choice of environment $e \in \Esp$---with the ultimate goal of generalizing to an unknown test dataset $p(x,y|e_{test})$.
Because optimal classifiers under standard loss functions can be realized via a conditional label distribution ($f^*(x) = \E[y|x]$), an invariant representation $\Phi(x)$ must satisfy the following \emph{Environment Invariance Constraint}: 
\begin{align}
\label{eq:irm}
    \E[y|\Phi(x) = h,e_1] = \E[y|\Phi(x) = h,e_2] \nonumber\\  \, \, \, \, \forall \, h \in \Zsp \, \, \, \, \forall \,  e_1, e_2 \in  \Esp.
\tag{\textsc{EIC}}
\end{align}
Intuitively, the representation $\Phi(x)$ encodes features of the input $x$ that induce the same conditional distribution over labels across each environment.
This is closely related to the notion of ``group sufficiency'' studied in the fairness literature \citep{liu2018implicit} (see Appendix \ref{sec:fairness_connections}).

Because trivial representations such as mapping all $x$ onto the same value may satisfy environment invariance, other objectives must be introduced to encourage the predictive utility of $\Phi$.
\citet{arjovsky2019invariant} propose IRM as a way to satisfy (\ref{eq:irm}) while achieving a good overall risk.
As a practical instantiation, the authors introduce IRMv1, a regularized objective enforcing simultaneous optimality of the same classifier $w \circ \Phi$ in all environments;\footnote{
$w \circ \Phi$ yields a classification decision via linear weighting on the representation features.}
here w.l.o.g. $w=\bar w$ is a constant scalar multiplier of 1.0 for each output dimension.
Denoting by $R^e = \E_{\pobs(x, y|e)}[\ell]$ the \mbox{per-environment} risk, the objective to be minimized is
\begin{equation}
    \label{eq:irmv1}
    C^{IRM}(\Phi) = 
    \sum_{e \in \Esp}
    R^e(\Phi)
    + \lambda ||
    \nabla_{\bar w}
    R^e(\bar w \circ \Phi)||.
    \tag{IRMv1}
\end{equation}

\paragraph{Robust Optimization}
Another approach at generalizing beyond the training distribution is robust optimization \citep{ben2009robust}, where one aims to minimize the worst-case loss for every subset of the training set, or other well-defined perturbation sets around the data \citep{duchi2021statistics, madry2017towards}.
Rather than optimizing a notion of invariance, Distributionally Robust Optimization (DRO) \citep{duchi2021statistics} seeks good performance for all nearby distributions by minimizing the worst-case loss: ${\max_q \E_{q}[\ell] \ \text{s.t.} \  D(q||p) < \epsilon}$, where $D$ denotes similarity between two distributions (e.g., $\chi^2$ divergence) and $\epsilon$ is a hyperparameter.
The objective can be computed as an expectation over $p$ via per-example importance weights ${\gamma_i = \frac{q(x_i, y_i)}{p(x_i, y_i)}}$.
GroupDRO operationalizes this principle by sharing importance weights across training examples, using environment labels to define relevant groups for this parameter sharing.
This can be expressed as an expected risk under a worst-case distribution over group proportions:
\[
C^{GroupDRO}(\Phi) = \max_g \E_{g(e)}[R^e(\Phi)]
\]
This is a promising approach towards tackling distribution shift with deep nets \citep{sagawa2019distributionally}, and we show in our experiments how environment inference enables application of GroupDRO to improve over standard learning without requiring group labels.

\paragraph{Limitations of Invariant Learning}
While the use of invariant learning to tackle domain generalization is still relatively nascent, several known limitations merit discussion.
IRM can provably find an invariant predictor that generalizes OOD, but only under restrictive assumptions, such as linearity of the data generative process and access to many environments \citep{arjovsky2019invariant}.
However, most benchmark datasets are in the non-linear regime; \citet{rosenfeld2020risks} demonstrated that for some non-linear datasets, the IRMv1 penalty term induces multiple optima, not all of which yield invariant predictors.
Nevertheless, IRM has found empirical success in some high dimensional non-linear classification tasks (e.g. CMNIST) using just a few environments \citep{arjovsky2019invariant,koh2020wilds}.
On the other hand, it was recently shown that, using careful and fair model selection strategies across a suite of image classification tasks, neither IRM nor other invariant learners consistently beat ERM in OOD generalization \citep{gulrajani2020search}.
This last study underscores the importance of \emph{model selection} in any domain generalization approach, which we discuss further below.

\section{Invariance Without Environment Labels}\label{sec:methods}
In this section we propose a novel invariant learning framework that does not require a priori domain/environment knowledge. 
This framework is useful in algorithmic fairness scenarios when demographic makeup is not directly observed; it is also applicable in standard machine learning settings when relevant environment information is either unavailable or not clearly identified.
In both cases, a method that sorts training examples $\mathcal{D}$ into environments that maximally separate the spurious features---i.e. inferring populations $\mathcal{D}_1 \cup \mathcal{D}_2=\mathcal{D}$---can facilitate effective
invariant learning.

\subsection{Environment Inference for Invariant Learning}\label{sec:eiil}

Our aim is to find environments that maximally violate the invariant learning
principle.
We can then evaluate the quality of these inferred environments
by utilizing them in an invariant learning method.
Our overall algorithm EIIL is a two-stage process: (1) Environment Inference (EI): infer the environment assignments; and (2) Invariant Learning (IL): run invariant learning given these assignments.

The primary goal of invariant-learning is to find features that are domain-invariant,
i.e, that reliably predict the true class regardless of the domain.
The EI phase aims to identify domains that help uncover these features.
This phase depends on a reference classifier
$\tilde \Phi$; which maps inputs $X$ to outputs $Y$, and defines a putative
set of invariant features.
This model could be
found using ERM on $\pobs(x,y)$, for example.
Environments are then derived that partition the mapping of the reference
model which maximally violate the invariance principle, i.e.,
where for the reference classifier the same feature vector is associated
with examples of different classes.
While any of the aforementioned invariant learning objectives can be
incorporated into the EI phase, the invariance principle or group-sufficiency---as expressed in (\ref{eq:irm})---is a natural fit, since it explicitly depends on learned feature representations $\Phi$.

To realize an EI phase focused on the invariance principle, we utilize the IRM objective (\ref{eq:irmv1}).
We begin by noting that the per-environment risk $R^e$ depends implicitly on the manual environment labels from the dataset.
For a given environment $e'$, we denote $\indicator{e_i = e'}$ as an indicator that example $i$ is assigned to that environment, and re-express the per-environment risk as:

\begin{align}
    R^{e}(\Phi) &= \frac{1}{\sum_{i'}\indicator{e_{i'} = e}
} \sum_i \indicator{e_i = e}
 \ell(\Phi(x_i), y_i)
\end{align}

Now we relax this risk measure to search over the space of environment assignments.
We replace the manual assignment indicator $\indicator{e_{i} = e'}$, with a probability distribution $\mathbf{q}_i(e'):=q(e'|x_i, y_i)$, representing a soft assignment of the $i$-th example to the $e'$-th environment.
To \emph{infer} environments, we optimize $q(e|x_i,y_i)$ so that it captures the worst-case environments for a fixed classifier $\Phi$.
This corresponds to maximizing w.r.t. $\mathbf{q}$ the following soft relaxation of the regularizer\footnote{
We omit the average risk term as we are focused on maximally violating (\ref{eq:irm}) regardless of the risk.
} from $C^{IRM}$:
\begin{align}
    C^{EI}(\Phi, \mathbf{q}) &= ||\nabla_{\bar w} \tilde R^e(\bar w \circ \Phi, \mathbf{q})||, \label{eq:ei-obj} \\
    \tilde R^e(\Phi, \mathbf{q}) &= \frac{1}{N} \sum_i \mathbf{q}_i(e) \ell(\Phi(x_i), y_i) \label{eq:soft-risk}
\end{align}
where $\tilde R^e$ represents a soft per-environment risk that can pass gradients to the environment assignments $\mathbf{q}$.
See Algorithm \ref{algo:eiil} in Appendix \ref{sec:pseudocode} for pseudocode.

To summarize, EIIL involves the following sequential\footnote{
We also tried jointly training $\Phi$ and $\mathbf{q}$ using alternating updates, as in GAN training, but did not find empirical benefits. 
This formulation introduces optimization and conceptual difficulties, e.g. ensuring that invariances apply to all environments discovered throughout learning.
} approach:
\begin{enumerate}
    \item Input \emph{reference model} $\tilde \Phi$;
    \item Fix $\Phi \leftarrow \tilde \Phi$ and optimize the EI objective to infer environments:  ${\mathbf{q}^* = \argmax_\mathbf{q} C^{EI}(\tilde \Phi, \mathbf{q})}$; 
    \item Fix $\mathbf{\tilde q} \leftarrow \mathbf{q}^*$ and optimize the IL objective to yield the new model: $\Phi^* = \argmin_\Phi C^{IL}(\Phi, \mathbf{\tilde q})$ 
\end{enumerate}

In our experiments we consider binary environments and parameterize the $\mathbf{q}$ as a vector of probabilities for each example in the training data.\footnote{Note that under this parameterization, when optimizing the inner loop with fixed $\Phi$ the number of parameters equals the number of data points (which is small relative
to standard neural net training).
We leave amortization of $q$ to future work.}
EIIL is applicable more broadly to any environment-based
invariant learning objective through the choice of $C^{IL}$ in Step 3.
We present experiments using $C^{IL}\in \{C^{IRM}, C^{GroupDRO}\}$, and leave a more complete exploration to future work.

\subsection{Analyzing the 
Inferred Environments}
\label{sec:analysis}
To characterize the ability of EIIL to generalize to unseen test data, we now examine the inductive bias for generalization provided by the reference model $\tilde \Phi$.
We state the main result here and defer the proofs to Appendix \ref{sec:proofs}.
Consider a dataset with some feature(s) $z$ which are spurious, and other(s) $v$ which are valuable/invariant/causal w.r.t. the label $y$.
Our proof considers binary features/labels and two environments, but the same argument extends to other cases.
Our goal is to find a model $\Phi$ whose representation $\Phi(v, z)$ is invariant w.r.t. $z$ and focuses solely on $v$.

\begin{proposition}\label{thm:1}
Consider environments that differ in the degree to which the label $y$ agrees with the spurious features $z$: $\mathbb{P}(\indicator{y=z}|e_1) \neq \mathbb{P}(\indicator{y=z}|e_2)$:
then a reference model $\tilde \Phi = \Phi_{Spurious}$ that is invariant to valuable features $v$ and solely focuses on spurious features $z$ maximally violates the invariance principle (\ref{eq:irm}).
Likewise, consider the case with fixed representation $\Phi$ that focuses on the spurious features: then a choice of environments that maximally violates (\ref{eq:irm}) is $e_1 = \{v,z,y|\indicator{y = z}\}$ and $e_2 = \{v,z,y|\indicator{y \neq z}\}$.
\end{proposition}

If environments are split according to agreement of $y$ and $z$,
then the constraint from (\ref{eq:irm}) is satisfied by a representation that ignores $z$: $\Phi(x) \perp z$.
Unfortunately this requires a priori knowledge of either the spurious feature $z$ or a reference model $\tilde \Phi = \Phi_{Spurious}$ that extracts it.
When the suboptimal solution
$\Phi_{Spurious}$ is not a priori known, it will sometimes be recovered directly from the training data; for example in CMNIST we find that $\Phi_{ERM}$ approximates $\Phi_{Color}$.
This allows EIIL to find environment partitions providing the starkest possible contrast for invariant learning.

Even if environment partitions are available, it may be possible to improve performance by inferring new partitions from scratch.
It can be shown (see Appendix \ref{sec:suboptimal_sufficiency}) that the environments provided in the {CMNIST} dataset \citep{arjovsky2019invariant} do not maximally violate (\ref{eq:irm}) for a reference model $\tilde \Phi = \Phi_{Color}$, and are thus not maximally informative for learning to ignore color.
Accordingly, EIIL improves test accuracy for IRM compared with the hand-crafted environments (Table \ref{tab:table_teaser}).

If $\tilde \Phi = \Phi_{ERM}$ focuses on a mix of $z$ and $v$, EIIL may still find environment partitions that enable effective invariant learning, as we find in the Waterbirds dataset, but they are not guaranteed to maximally violate (\ref{eq:irm}). 

\subsection{Binned Environment Invariance}\label{sec:binned-ei}

We can derive a heuristic algorithm for EI that maximizes violations of the invariance principle by stratifying examples into discrete bins (i.e. confidence bins for 1-D representations), then sorting them into environments within each bin.
This algorithm provides insight into both the EI task and the relationship between the IRMv1 regularizer and the invariance principle.
We define bins in the space of the learned representation $\Phi(x)$,
indexed by $b$; $s_{ib}$ indicates whether example $i$ is in bin $b$.
The intuition behind the algorithm is that a simple approach can separate the examples in a bin to achieve the maximal value of the (\ref{eq:irm}).

The degree to which the environment assignments violate (\ref{eq:irm}) can be expressed as follows, which can then be approximated in terms of the bins:
{\small
\begin{eqnarray}
\Delta \rm{EIC} & = & (E[y|\Phi(x),e_1] - E[y|\Phi(x),e_2])^2 \label{eq:softEIC} \nonumber\\ 
& \approx & \sum_b (\sum_i s_{ib} y_i \mathbf{q}_i(e=e_1) - \sum_i s_{ib} y_i \mathbf{q}_i(e=e_2))^2 \nonumber\label{eq:binEIC}
\end{eqnarray}
}
Inspection of this objective leads to a simple algorithm: assign all the $y=1$ examples to one
environment, and $y=-1$ examples to the other. This results in the
expected values of $y$ equal to $\pm 1$, which achieves the maximum possible value of $\Delta \rm{EIC}$ per bin.\footnote{For very confident reference models, where few confidence bins are populated, splitting based on $y$ relates to partitioning the error cases. This splitting strategy is not the only possible solution, as multiple global optima exist.
}

This binning leads to an important insight into the relationship 
between the IRMv1 regularizer and (\ref{eq:irm}). Despite the analysis in \citet{arjovsky2019invariant}, this link is not completely
clear \citep{kamath2021does, rosenfeld2020risks}.
However, in the situation considered here, with binary classes, we can use this binning approach to show a tight link between the two objectives:
finding an environment assignment that maximizes the
violation of our softened IRMv1 regularizer (Equation \ref{eq:ei-obj}) also maximizes the violation
of the softened Environment Invariance Constraint ($\Delta \rm{EIC}$); see Appendix \ref{sec:majmin} for the proof.
This binning approach highlights the dependence on the reference model, as the bins are defined in its learned $\Phi$ space; the reference model also played a key role in the analysis above. We analyze it empirically in Section \ref{sec:secondary-results}.

%%%%%%%%%%%%%%%%%%%%%%%%%%%%%%%%%%%%%%%%%%%%%%%%%%%%%%%%%%%%
% related works table
%%%%%%%%%%%%%%%%%%%%%%%%%%%%%%%%%%%%%%%%%%%%%%%%%%%%%%%%%%%%
\begin{table*}[th]
\centering
\resizebox{\textwidth}{!}{
\begin{tabular}{lccc}
\toprule
{
Statistic to match/optimize
} & $e$ known? &DG method & Fairness method \\
\midrule
\midrule
match $\E[\ell|e] \ \forall e$ & yes & REx \citep{krueger2020out}, & CVaR Fairness \citep{williamson2019fairness} \\
\midrule
$\min \max_e\E[\ell|e]$
             & yes & Group DRO \citep{sagawa2019distributionally} & \\
\midrule
$\min \max_q\E_q[\ell]$ & no & DRO \citep{duchi2021statistics} & Fairness without Demographics\\
& & & \citep{hashimoto2018fairness,lahoti2020fairness} \\
\midrule
match $\E[y|\Phi(x), e] \ \forall \ e$ & yes & IRM \citep{arjovsky2019invariant} & Group Sufficiency \\
& & & \citep{chouldechova2017fair,liu2018implicit}\\
\midrule
match $\E[y|\Phi(x),e] \ \forall \ e$ & no & \textbf{EIIL (ours)} & \textbf{EIIL (ours)}\\
\midrule
match $\E[\hat y|\Phi(x),e, y=y'] \ \forall \ e$ & yes & C-DANN \citep{li2018deep} & Equalized Odds \citep{hardt2016equality}\\
& & PGI \citep{ahmed2021systematic} & \\

\midrule
match $\Big|\E[y|S(x),e]-\E[\hat y(x)| S(x), e]\Big| \ \forall \ e$ & no 
& & Multicalibration \citep{hebert2017calibration}\\
\midrule
match $\Big|\E[y|e]-\E[\hat y(x)|e]\Big| \ \forall \ e$ & no & & Multiaccuracy \citep{kim2019multiaccuracy}\\
\midrule
match $\Big|\E[y \neq \hat y(x)|y=1,e]\Big| \ \forall \ e$& no & & Fairness Gerrymandering \citep{kearns2018preventing}\\

\bottomrule
\end{tabular}
}
\caption{Domain Generalization (DG) and Fairness methods can be understood as matching or optimizing some statistic across conditioning variable $e$, representing ``environment'' or ``domains'' in DG and ``sensitive'' group membership in the Fairness.
$\Phi$ and $S$ are learned vector and scalar functions of the inputs, respectively. 
}
\label{tab:relating_dom_gen_to_fairness}
\end{table*}
%%%%%%%%%%%%%%%%%%%%%%%%%%%%%%%%%%%%%%%%%%%%%%%%%%%%%%%%%%%%

\section{Related Work}\label{sec:related-work}

\paragraph{Domain adaptation and generalization}

Beyond the methods discussed above, a variety of recent works have approached the domain generalization problem from the lens of learning invariances in the training data.
Adversarial training is a popular approach for learning representations invariant \citep{zhang2017aspect,hoffman2018cycada,ganin2016domain} or conditionally invariant \citep{li2018deep} to the environment.
However, this approach has limitations in settings where distribution shift affects the marginal distribution over labels \citep{zhao2019learning}.

\citet{arjovsky2019invariant} proposed IRM to mitigate the effect of test-time label shift, which was inspired by applications of causal inference to select invariant features  \citep{peters2016causal}.
\citet{krueger2020out} proposed the related Risk Extrapolation (REx) principle, which dictates a stronger preference to exactly equalize $R^e \ \forall \ e$ (e.g. by penalizing variance across $e$ as in their practical algorithm V-REx), which is shown to improve generalization in several settings.\footnote{
Analogous to V-REx, \citet{williamson2019fairness} adapt Conditional Variance at Risk (CVaR) \citep{rockafellar2002conditional} to equalize risk across demographic groups.
}

Recently, \citet{ahmed2021systematic} proposed a new invariance regularizer based on matching class-conditioned average predictive distributions across environments, which we note is closely related to the equalized odds criterion commonly used in fair classification \citep{hardt2016equality}.
Moreover, they deploy this training on top of environments inferred by our EI method, showing that the overall EIIL approach can effectively handle ``systematic'' generalization \citep{bahdanau2018systematic} on a semi-synthetic foreground/background task similar to the Waterbirds dataset that we study.

Several large-scale benchmarks have recently been proposed to highlight difficult open problems in this field, including the use of real-world data \citep{koh2020wilds}, handling subpopulation shift \citep{santurkar2020breeds}, and model selection \citep{gulrajani2020search}.

\paragraph{Leveraging a reference classifier}
A number of methods have recently been proposed that improve performance by exploiting the mistakes of a pre-trained auxiliary model, as we do when inferring environments for the invariant learner using $\tilde \Phi$.
\citet{nam2020learning} jointly train a ``biased'' model $f_B$ and a ``debiased'' model $f_D$, where the relative cross-entropy losses of $f_B$ and $f_D$ on each training example determine their importance weights in the overall training objective for $f_D$.
\citet{sohoni2020no} infer a different set of ``hidden subclasses`` for each class label $y \in \Ysp$, Subclasses computed in this way are then used as group labels for training a GroupDRO model, so the overall two-step process corresponds to certain choices of EI and IL objectives. 

\citet{liu2021just} and \citet{dagaev2021too} concurrently proposed to compute importance weights for the primary model using an ERM reference, which can be seen as a form of distributionally robust optimization where the worst-case distribution only updates once.
\citet{dagaev2021too} use the confidence of the reference model to assign importance weights to each training example.
\citet{liu2021just} split the training examples into two disjoint groups based on the errors of ERM, akin to our EI step, with the per-group importance weights treated as a hyperparameter for model selection (which requires a subgroup-labeled validation set).
We note that the implementation of EIIL using binning, discussed in Section \ref{sec:binned-ei}, can also realize an error splitting behavior. 
%when the reference classifier is very confident.
In this case, both methods use the same disjoint groups of training examples towards slightly different ends: we train an invariant learner, whereas \citet{liu2021just} train a cross-entropy classifier with fixed per-group importance weights.

\paragraph{Algorithmic fairness}

Our work draws inspiration from a rich body of recent work on learning fair classifiers in the absence of demographic labels \citep{hebert2017calibration,kearns2018preventing,hashimoto2018fairness,kim2019multiaccuracy,lahoti2020fairness}.
Generally speaking, these works seek a model that performs well for group assignments that are the worst case according to some fairness criterion.
Table \ref{tab:relating_dom_gen_to_fairness} enumerates
several of these criteria, and draw analogies to domain generalization methods that match or optimize similar statistics.\footnote{
We refer the interested reader to Appendix \ref{sec:fairness_connections} for a more in-depth discussion of the relationships between domain generalization and fairness methods.
}
Environment inference serves a similar purpose for our method, but with a slightly different motivation: rather than learn an fair model in an online way that provides favorable in-distribution predictions, we learn discrete data partitions as an intermediary step, which enables use of invariant learning methods to tackle distribution shift.

Adversarially Reweighted Learning (ARL) \citep{lahoti2020fairness} is most closely related to ours, since they emphasize subpopulation shift as a key motivation.
Whereas ARL uses a DRO objective that prioritizes stability in the loss space, we explore environment inference to encourage invariance in the learned representation space.
We see these as complementary approaches that are each suited to different types of distribution shift, as we discuss in the experiments.

\section{Experiments}\label{sec:experiments}
For lack of space we defer a proof-of-concept synthetic regression experiment to Appendix \ref{sec:synthetic-regression}.
We proceed by describing the remaining datasets under study in Section \ref{sec:datasets}.
We then present the main results measuring the ability of EIIL to handle distribution shift in Section \ref{sec:primary-results}, and offer a more detailed analysis of the EIIL solution and its dependence on the reference model in Section \ref{sec:secondary-results}.
See \mbox{\url{https://github.com/ecreager/eiil}} for code.

\paragraph{Model selection}
Tuning hyperparameters when train and test distributions differ is a difficult open problem \citep{krueger2020out,gulrajani2020search}.
Where possible, we reuse effective hyperparameters for IRM and GroupDRO found by previous authors.
Because these works allowed limited validation samples for hyperparameter tuning (all baseline methods benefit fairly from this strategy), these results represent an optimistic view on the ability for invariant learning.
As discussed above, the choice of reference classifier is of crucial importance when deploying EIIL; if worst-group performance can be measured on a validation set, this could be used to tune the hyperparameters of the reference model (i.e. model selection subsumes reference model selection).
See Appendix \ref{sec:experimental_details} for further discussion.

\subsection{Datasets}\label{sec:datasets}
\paragraph{CMNIST}
\newcommand{\lblnoise}{\theta_y}
CMNIST is a noisy digit recognition task\footnote{MNIST digits are grouped into $\{0,1,2,3,4\}$ and $\{5,6,7,8,9\}$ so the CMNIST target label $y$ is binary.}
where color is a spurious feature that correlates with the label at train time but anti-correlates at test time, with the correlation strength at train time varying across environments \citep{arjovsky2019invariant}.
In particular, the two training environments have $Corr(color, label) \in \{0.8,0.9\}$ while the test environment has $Corr(color, label)=0.1$.
Crucially, label noise is applied by flipping $y$ with probability $\lblnoise = 0.25$.
This implies that shape (the invariant feature) is marginally less reliable than color in the training data, so ERM ignores shape to focus on color and suffers from below-chance test performance.

\paragraph{Waterbirds}

To evaluate whether EIIL can infer useful environments in a more challenging setting with high-dimensional images, we turn to the Waterbirds dataset \citep{sagawa2019distributionally}.
Waterbirds is a composite dataset that combines $4,795$ bird images from the CUB dataset \citep{wah2011caltech} with background images from the Places dataset \citep{zhou2017places}.
It examines the proposition (which frequently motivates invariant learning approaches) that modern networks often learn spurious background features (e.g. green grass in pictures of cows) that are predictive of the label at train time but fail to generalize in new contexts \citep{beery2018recognition,geirhos2020shortcut}.
The target labels are two classes of birds---``landbirds'' and ``waterbirds'' respectively coming from dry or wet habitats---superimposed on land and water backgrounds.
At training time, landbirds and waterbirds are most frequently paired with land and water backgrounds, respectively, but at test time the 4 subgroup combinations are uniformly sampled.
To mitigate failure under distribution shift, a robust representation should learn primarily features of the bird itself, since these are invariant, rather than features of the background.
Beyond the increase in dimensionality, this task  differs from CMNIST in that the ERM solution does not fail catastrophically at test time, and in fact can achieve 97.3\% average accuracy.
However, because ERM optimizes average loss, it suffers in performance on the worst-case subgroup (waterbirds on land, which has only $56$ training examples).

\paragraph{Adult-Confounded}
To assess the ability of EIIL to address 
worst-case group performance without group labels, we construct a variant of the UCI Adult dataset,\footnote{\url{https://archive.ics.uci.edu/ml/datasets/adult}}
which comprises $48,842$ census records collected from the USA in 1994. 
The task commonly used as an algorithmic fairness benchmark is to predict a binarized income indicator (thresholded at $\$50,000$) as the target label, possibly considering sensitive attributes such as age, sex, and race.

\citet{lahoti2020fairness} demonstrate the benefit of per-example loss reweighting on Adult using their method ARL to improve predictive performance for undersampled subgroups.
Following \citet{lahoti2020fairness}, we consider the effect of four sensitive \emph{subgroups}---defined by composing binarized race and sex labels---on model performance, assuming the model does not know a priori which features are sensitive.
However, we focus on a distinct generalization problem where a pernicious dataset bias confounds the training data, making subgroup membership predictive of the label on the training data. 
At test time these correlations are reversed, so a predictor that infers subgroup membership to make predictions will perform poorly at test time (see Appendix \ref{sec:dataset_details} for  details).
This large test-time distribution shift can be understood as a controlled \emph{audit} to determine if the classifier uses subgroup information to predict the target label.
We call our dataset variant Adult-Confounded.

\paragraph{CivilComments-WILDS}

We apply EIIL to a large and challenging comment toxicity prediction task with important fairness implications \citep{borkan2019nuanced}, where ERM performs poorly on comments associated with certain identity groups.
We follow the procedure and data splits of \citet{koh2020wilds} to finetune DistilBERT embeddings \citep{sanh2019distilbert}.
EIIL uses an ERM reference classifier and its inferred environments are fed to a GroupDRO invariant learner.
Because the large training set ($N_{train}=269,038$) increases the convergence time for gradient-based EI, we deploy the binning heuristic discussed in Section \ref{sec:binned-ei}, which in this instance finds environments that correspond to the error and non-error cases of the reference classifier.
While ERM and EIIL do not have access to the sensitive group labels, we note that worst-group validation accuracy is used to tune hyperparameters for all methods.
See Appendix \ref{sec:experimental_details} for details.
We also compare against a GroupDRO (oracle) learner that has access to group labels.

\subsection{Results}\label{sec:primary-results}

\paragraph{CMNIST}

%%%%%%%%%%%%%%%%%%%%%%%%%%%%%%%%%%%%%%%%%%%%%%%%%%%%%%
% Main CMNIST result
%%%%%%%%%%%%%%%%%%%%%%%%%%%%%%%%%%%%%%%%%%%%%%%%%%%%%%
\definecolor{myred}{RGB}{215,48,39}
\definecolor{mygreen}{RGB}{26,152,80}
\newcommand{\cmark}{\textcolor{mygreen}{\ding{51}}}
\newcommand{\xmark}{\textcolor{myred}{\ding{55}}}
\begin{table}[h]
\centering
\begin{tabular}{cccc}
\toprule
{Method} & Handcrafted &      Train &       Test \\
 & Environments &      &        \\
\midrule
ERM       & \xmark &  \textbf{86.3 $\pm$ 0.1} &  13.8 $\pm$ 0.6 \\
IRM  & \cmark & 71.1 $\pm$ 0.8 &  65.5 $\pm$ 2.3 \\
\textbf{EIIL}
& \xmark   &  73.7 $\pm$ 0.5 &  \textbf{68.4 $\pm$ 2.7} \\
\bottomrule
\end{tabular}
\caption{
Accuracy (\%) on CMNIST, a digit classification task where color is a spurious feature correlated with the label during training but anti-correlated at test time.
EIIL exceeds test-time performance of IRM \emph{without} knowledge of pre-specified environment labels, by instead finding worst-case environments using aggregated data and a reference classifier.
}
\label{tab:table_teaser}
\end{table}
%%%%%%%%%%%%%%%%%%%%%%%%%%%%%%%%%%%%%%%%%%%%%%%%%%%%%%

IRM was previously shown to learn an invariant representation on this dataset, allowing it to generalize relatively well to the held-out test set whereas ERM fails dramatically \citep{arjovsky2019invariant}.
It is worth noting that label noise makes the problem challenging, so even an oracle classifier can achieve at most 75\% test accuracy on this binary classification task.
To realize EIIL in our experiments, we discard the environment labels, and run the procedure described in Section \ref{sec:eiil} with ERM as the reference model and IRM as the invariant learner used in the final stage.
We find that EIIL's environment labels are very effective for invariant learning, ultimately \emph{outperforming} standard IRM using the environment labels provided in the dataset 
(Table \ref{tab:table_teaser}).
This suggests that in this case the EIIL solution approaches the \emph{maximally informative} set of environments discussed in Proposition \ref{thm:1}.

\paragraph{Waterbirds}

\citet{sagawa2019distributionally} demonstrated that ERM suffers from poor worst-group performance on this dataset, and that GroupDRO can mitigate this performance gap if group labels are available.
In this dataset, group labels should be considered as oracle information, meaning that the relevant baseline for EIIL is standard ERM.
The main contribution of \citet{sagawa2019distributionally} was to show \emph{how} deep nets can be optimized for the GroupDRO objective using their online algorithm that adaptively adjusts per-group importance weights.
In our experiment, we combine this insight with our EIIL framework to show that distributionally robust neural nets can be realized without access to oracle information.
We follow the same basic procedure as above,\footnote{For this dataset, environment inference worked better with reference models that were not fully trained. We suspect this is because ERM focuses on the easy-to-compute features like background color early in training, precisely the type of bias EIIL can exploit to learn informative environments.} in this case using GroupDRO as the downstream invariant learner for which EIIL's inferred labels will be used.

%%%%%%%%%%%%%%%%%%%%%%%%%%%%%%%%%%%%%%%%%%%%%%%%%%%%%
% waterbirds main table
%%%%%%%%%%%%%%%%%%%%%%%%%%%%%%%%%%%%%%%%%%%%%%%%%%%%%
\begin{table}[ht]
\centering
\resizebox{.48\textwidth}{!}{
\begin{tabular}{lccc}
\toprule
{Method} &  Train (avg) & Test (avg) & Test (worst group) \\
\midrule
ERM & 100.0 &  \textbf{97.3} &  60.3 \\
EIIL  & 99.6 &  96.9 & \textbf{78.7} \\
\midrule
GroupDRO (oracle) & 99.1  & 96.6 &  84.6 \\
\bottomrule
\end{tabular}
}
\caption{Accuracy (\%) on the Waterbirds dataset.
EIIL strongly outperforms ERM on worst-group performance, approaching the performance of the GroupDRO algorithm proposed by \citet{sagawa2019distributionally}, which requires oracle access to group labels.
In this experiment we feed environments inferred  by EIIL into a GroupDRO learner.
}
\label{tab:waterbirds_results}
\end{table}
%%%%%%%%%%%%%%%%%%%%%%%%%%%%%%%%%%%%%%%%%%%%%%%%%%%%%%

%%%%%%%%%%%%%%%%%%%%%%%%%%%%%%%%%%%%%%%%%%%%%%%%%%%%%%
% waterbirds inferred envs figure
%%%%%%%%%%%%%%%%%%%%%%%%%%%%%%%%%%%%%%%%%%%%%%%%%%%%%%
\begin{figure}[thb]
\centering
\includegraphics[width=.45\textwidth]{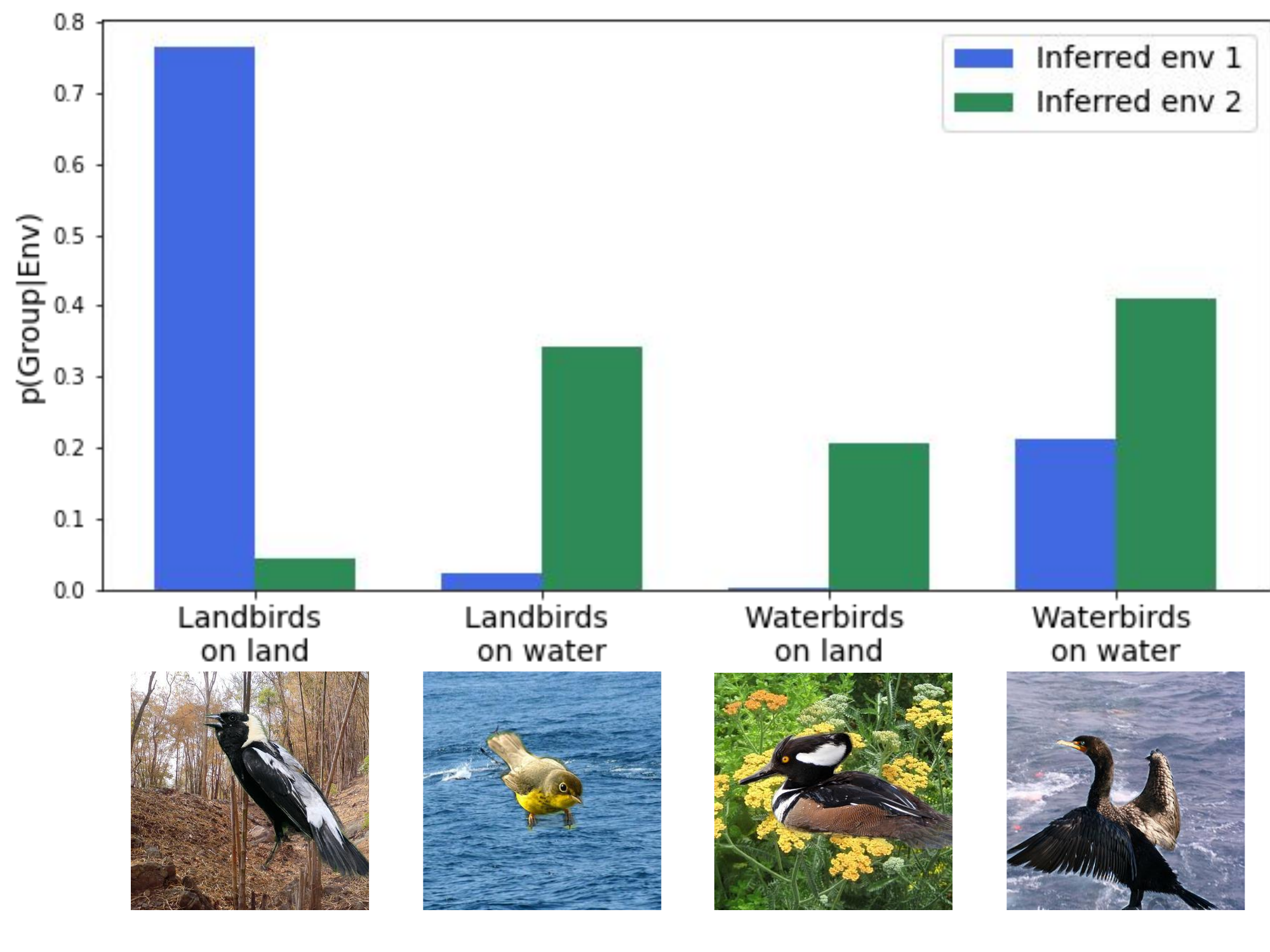}
\caption[]{After using EIIL to directly infer two environments from the Waterbirds dataset, we examine the proportion of each subgroup (available in the original dataset but not used by EIIL) present in the inferred environment.
}
\label{fig:waterbirds_inferred_envs}
\end{figure}
%%%%%%%%%%%%%%%%%%%%%%%%%%%%%%%%%%%%%%%%%%%%%%%%%%%%%%

EIIL is significantly more robust than the ERM baseline (Table \ref{tab:waterbirds_results}), raising worst-group test accuracy by 18\% with only a 1\% drop in average accuracy.
In Figure \ref{fig:waterbirds_inferred_envs} we plot the distribution of subgroups for each inferred environment, showing that the minority subgroups (landbirds on water and waterbirds on land) are mostly organized in the same inferred environment.
This suggests the possibility of leveraging environment inference for interpretability to automatically discover a model's performance discrepancies on subgroups, which we leave for future work.

\paragraph{Adult-Confounded}

\begin{table}[b!]
\centering
%%%%%%%%%%%%%%%%%%%%%%%%%%%%%%%%%%%%%%%%%%%%%%%%%%%%%%
% UCIAdult-Confounded main table
%%%%%%%%%%%%%%%%%%%%%%%%%%%%%%%%%%%%%%%%%%%%%%%%%%%%%%
\begin{tabular}{lll}
\toprule
{Method} &      Train accs &       Test accs \\
\midrule
ERM &  \textbf{92.7 $\pm$ 0.5} &  31.1 $\pm$ 4.4 \\
ARL \citep{lahoti2020fairness} &  72.1 $\pm$ 3.6 &  61.3 $\pm$ 1.7 \\
EIIL &  69.7 $\pm$ 1.6 &  \textbf{78.8 $\pm$ 1.4} \\
\bottomrule
\end{tabular}
\caption{
Accuracy on Adult-Confounded, a variant of the UCI Adult dataset where some sensitive subgroups correlate to the label at train time, and this correlation pattern is reversed at test time.
}
\label{tab:confounded_adult}
\end{table}
%%%%%%%%%%%%%%%%%%%%%%%%%%%%%%%%%%%%%%%%%%%%%%%%%%%%%%

%%%%%%%%%%%%%%%%%%%%%%%%%%%%%%%%%%%%%%%%%%%%%%%%%%%%%%
% UCIAdult-Confounded calibration plots
%%%%%%%%%%%%%%%%%%%%%%%%%%%%%%%%%%%%%%%%%%%%%%%%%%%%%%
\begin{figure}[hbt]
\newcommand{\mywidth}{.23\textwidth}
\centering
\subfigure{\includegraphics[width=\mywidth]{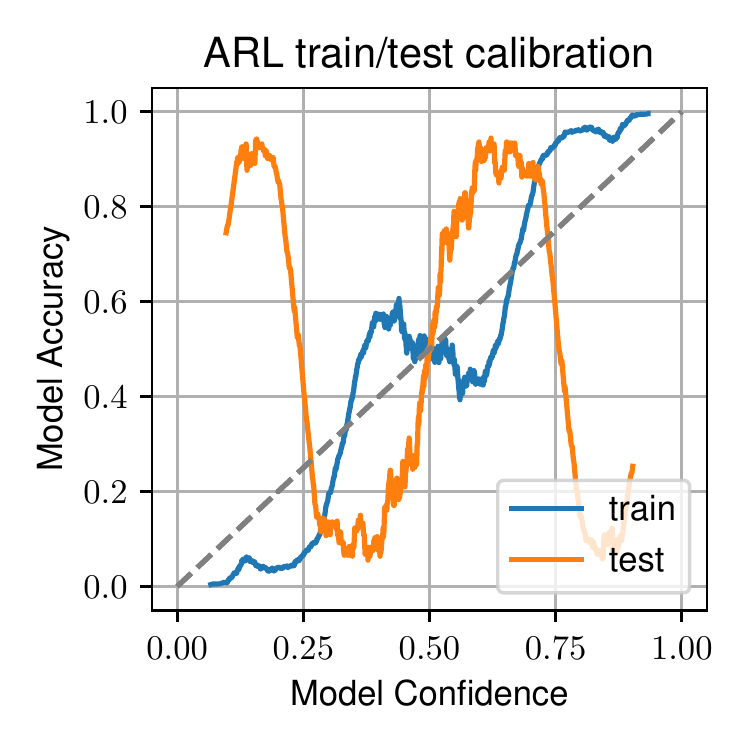}}
\subfigure{\includegraphics[width=\mywidth]{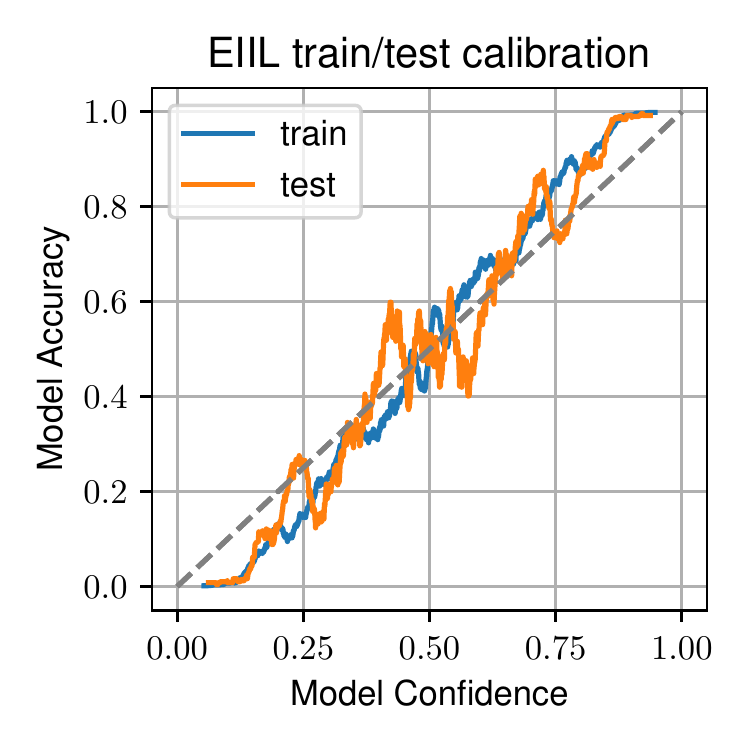}}
\caption{
By inferring environments that maximally violate the invariance principle, and then applying invariant learning across the inferred environments, EIIL finds a solution that is well calibrated on the test set (right), compared with ARL (left).
}
\label{fig:group_sufficiency}
\end{figure}
%%%%%%%%%%%%%%%%%%%%%%%%%%%%%%%%%%%%%%%%%%%%%%%%%%%%%%

Using EIIL---to first infer worst-case environments then ensure invariance across them---performs favorably on the audit test set, compared with ARL and an ERM baseline (Table \ref{tab:confounded_adult}).
We also find that, without access to sensitive group information, the solution found by EIIL achieves significantly better calibration on the test distribution (Figure \ref{fig:group_sufficiency}).
Because the train and test distributions differ based on the correlation pattern of small subgroups to the target label, this suggests that EIIL has achieved favorable group sufficiency \citep{liu2018implicit} in this setting.
See Appendix \ref{sec:ablation} for a discussion of this point, as well as an ablation showing that all components of the EIIL approach are needed to achieve the best performance.

\paragraph{CivilComments-WILDS}

%%%%%%%%%%%%%%%%%%%%%%%%%%%%%%%%%%%%%%%%
% CivilComments table
\begin{table}[b]
\vspace{-.1cm}
\centering
\resizebox{0.5\textwidth}{!}{%
\begin{tabular}{lcccccc}
\toprule
{Method} &        Train (avg) & Test (avg) & Test (worst group) \\
\midrule
ERM  &  96.0 $\pm$ 1.5 &   \textbf{92.0 $\pm$ 0.4} &  61.6 $\pm$ 1.3 \\
EIIL &  97.0 $\pm$ 0.8 &   90.5 $\pm$ 0.2 &  \textbf{67.0 $\pm$ 2.4} \\
\midrule
GroupDRO (oracle) &  93.6 $\pm$ 1.3 &  89.0 $\pm$ 0.3 &  69.8 $\pm$ 2.4 \\
\bottomrule
\end{tabular}
}
\caption{
EIIL improves worst-group accuracy in the CivilComments-WILDS toxicity prediction task,
without access to group labels.
}
\end{table}
%%%%%%%%%%%%%%%%%%%%%%%%%%%%%%%%%%%%%%%%

Without knowledge of which comments are associated with which groups, EIIL improves worst-group accuracy over ERM with only a modest cost in average accuracy, approaching the oracle GroupDRO solution (which requires group labels).

\subsection{Influence of the reference model}\label{sec:secondary-results}
As discussed in Section \ref{sec:analysis}, the ability of EIIL to find useful environments---partitions yielding an invariant representation when used by an invariant learner---depends on its ability to exploit variation in the predictive distribution of a reference classifier.
Here we study the influence of the reference classifier on the final EIIL solution.
We return to the CMNIST dataset, which provides a controlled sampling setup where particular ERM solutions can be induced to serve as reference for EIIL.
In Appendix \ref{sec:synthetic-regression}, we discuss a similar experiment in a synthetic regression setting.

EIIL was shown to outperform IRM \emph{without access to environment labels} in the standard CMNIST dataset (Table \ref{tab:table_teaser}), which has label noise of $\lblnoise=0.25$.
Because $Corr(color, label$) is 0.85 (on average) for the train set, this amount of label noise implies that color is the most predictive feature on aggregated training set (although its predictive power varies across environments).
ERM, even with access to infinite data, will focus on color given this amount of label noise to achieve an average train accuracy of 85\%.
However we can implicitly control the ERM solution $\Phi_{ERM}$ by tuning $\theta_y$, an insight that we use to study the dependence of EIIL on the reference model $\tilde \Phi = \Phi_{ERM}$.

%%%%%%%%%%%%%%%%%%%%%%%%%%%%%%%%%%%%%%%%%%%%%%%%%%%%%%
% follow-up CMNIST plot - analysis of ref model
%%%%%%%%%%%%%%%%%%%%%%%%%%%%%%%%%%%%%%%%%%%%%%%%%%%%%%
\begin{figure}[ht]
\centering
\subfigure[Train accuracy.]{
\includegraphics[width=.22\textwidth]{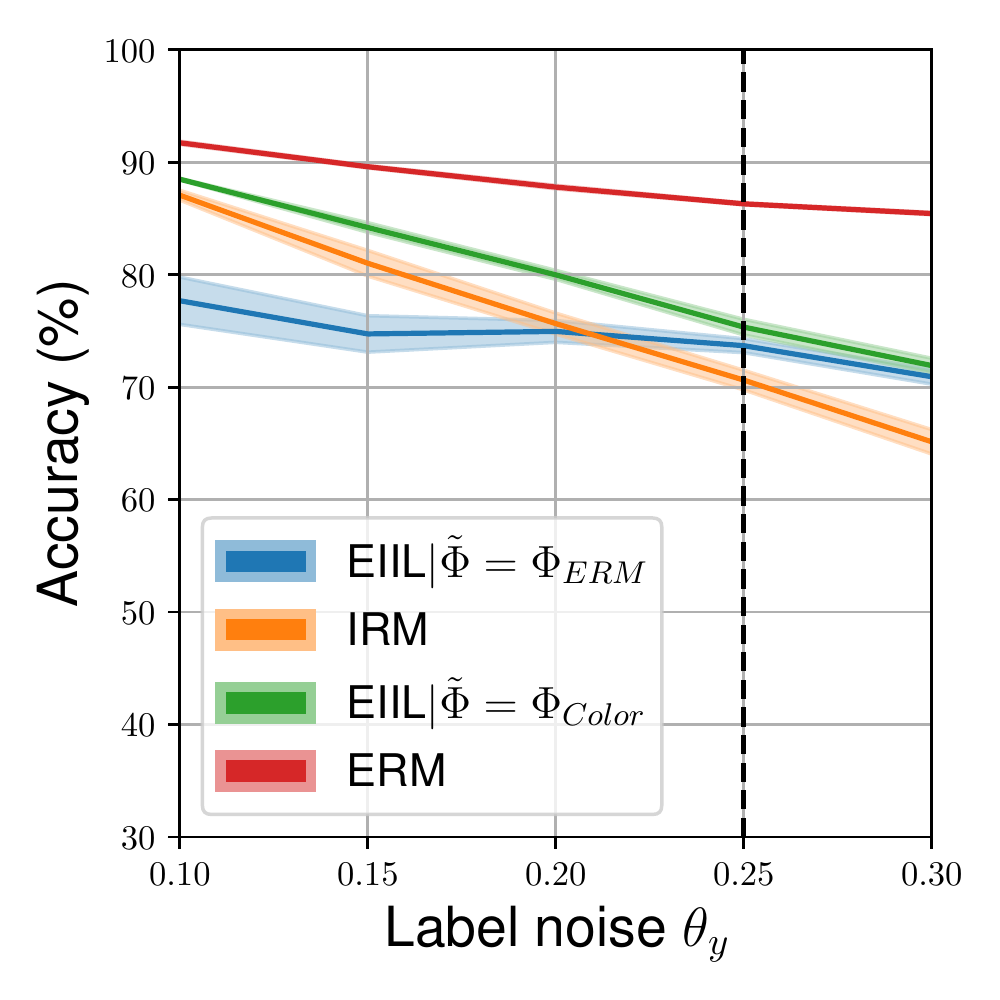}
\label{fig:label_noise_sweep_test_accs}
}
\subfigure[Test accuracy]{
\includegraphics[width=.22\textwidth]{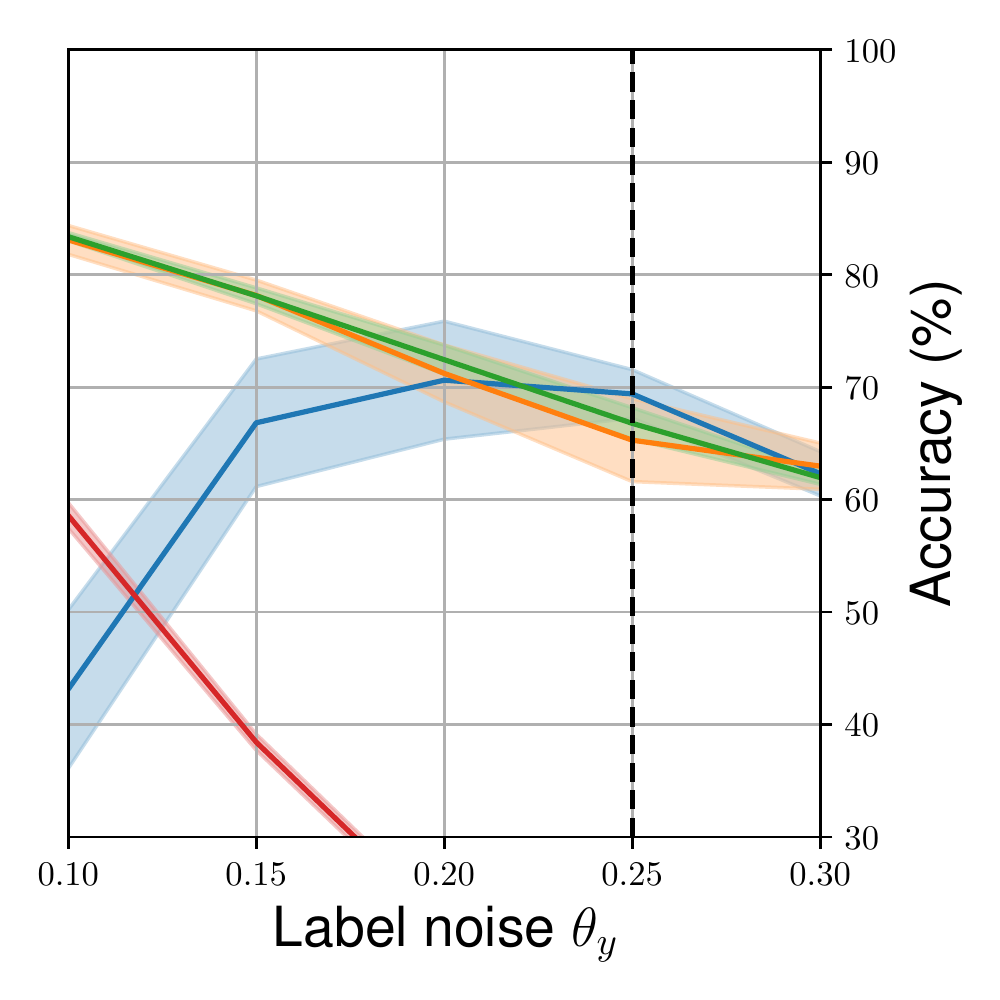}
\label{fig:label_noise_sweep_train_accs}
}
\vspace{-3mm}
\caption[]{
CMNIST with varying label noise $\lblnoise$.
Under high label noise (${\lblnoise > .2}$), where the spurious feature color correlates to label \emph{more} than shape on the train data, EIIL matches or exceeds the test performance of IRM \emph{without relying on hand-crafted environments}.
Under medium label noise (${.1 < \lblnoise < .2}$), EIIL is worse than IRM but better than ERM, the logical approach if environments are not available.
Under low label noise (${\lblnoise < .1}$), where color is \emph{less} predictive than shape at train time, ERM  performs well and EIIL  fails.
The vertical dashed black line indicates the default setting of $\theta_y=0.25$, which we report in Table
\ref{tab:table_teaser}.
}
\label{fig:label_noise_sweep}
\end{figure}
%%%%%%%%%%%%%%%%%%%%%%%%%%%%%%%%%%%%%%%%%%%%%%%%%%%%%%

Figure \ref{fig:label_noise_sweep} shows the results of our study.
We find that EIIL generalizes better than IRM  with sufficiently high label noise $\lblnoise > .2$, but generalizes poorly under low label noise.
This is precisely because ERM learns the color feature under high label noise, and the shape feature under low label noise.
We verify this conclusion by evaluating EIIL when $\tilde \Phi = \Phi_{Color}$, i.e. a hand-coded color-based predictor as reference, which does well across all settings of $\lblnoise$.

We saw in the Waterbirds experiment that it is not a strict requirement that ERM fail completely in order for EIIL to succeed.
However, this controlled study highlights the importance of the reference model in the ability of EIIL to find environments that emphasize the right invariances, which leaves open the question of how to effectively choose a reference model for EIIL in general.
One possible way forward is by using validation data that captures the \emph{kind} of distribution shift we expect at test time, without exactly producing the test distribution, e.g. as in the WILDS benchmark \citep{koh2020wilds}.
In this case we could choose to run EIIL with a reference model that exhibits a large generalization gap between the training and validation distributions.

\section{Conclusion}
We introduced EIIL, a new method that infers environment partitions of aggregated training data for invariant learning. Without access to environment labels, EIIL can outperform or approach invariant learning methods that require environment labels. EIIL has implications for domain generalization and fairness alike, because in both cases it can be hard to specify meaningful environments or sensitive subgroups.

\section*{Acknowledgements}
We are grateful to David Madras, Robert Adragna, Silviu Pitis, Will Grathwohl, Jesse Bettencourt, and Eleni Triantafillou for their feedback on this manuscript.
Resources used in preparing this research were provided, in part, by the Province of Ontario, the Government of Canada through \mbox{CIFAR}, and companies sponsoring the Vector Institute (\mbox{\url{www.vectorinstitute.ai/partners}}).

\bibliography{refs}
\bibliographystyle{icml2021}

\appendix
\onecolumn

\section{Environment Inference Psuedocode}\label{sec:pseudocode}
\icmltitlerunning{Environment Inference for Invariant Learning - Supplemental Material}

%%%%%%%%%%%%%%%%%%%%%%%%%%%%%%%%%%%%%%%%%%%%%%%%%%%%%%%%%%%%
% algo box
%%%%%%%%%%%%%%%%%%%%%%%%%%%%%%%%%%%%%%%%%%%%%%%%%%%%%%%%%%%%
\begin{algorithm}[tb]
   \caption{Pseudocode for environment inference (EI) with the invariance principle (realized via relaxed IRMv1 penalty) as the EI objective.}
   \label{alg:example}
\begin{algorithmic}
   \STATE {\bfseries Input:} Reference model $\Phi$, dataset $\mathcal{D} = \{x_i, y_i\}$, loss $\ell$, duration $N_{steps}$
   \STATE {\bfseries Output:} Worst case data splits $\mathcal{D}_1$, $\mathcal{D}_2$ for use with an invariant learner.
   \vspace{.2cm}
   \STATE \textbf{def} $\tilde R^e(\Phi, \mathbf{q})$:
   \STATE \ \ \textbf{return} $\frac{1}{N} \sum_i \mathbf{q}_i(e) \ell(\Phi(x_i), y_i)$ \hfill\COMMENT{Equation \ref{eq:soft-risk}}
   \vspace{.2cm}
   \STATE Randomly init. $\mathbf{q} \in [0, 1]^N$ environment posterior ($\mathbf{q}_i(e) := q(e|x_i, y_i)$)
   \FOR{$n \in 1 \ldots N_{steps}$}
     \STATE $SoftVari = \sum_{e\in\{1,2\}}||\nabla_{\bar w} \tilde R^e(\bar w \circ \Phi, \mathbf{q})||$
     \hfill \COMMENT{Aggregate reference model variances across soft envs}
     \STATE $Loss = -1 \cdot SoftVari$
     \hfill \COMMENT{Maximize the EI objective by minimizing this loss}
     \STATE $\mathbf{q} \leftarrow OptimUpdate(\mathbf{q}, \nabla_{\mathbf{q}} Loss)$
   \ENDFOR
   \STATE $\hat{ \mathbf{q} }\sim {Bernoulli}(\mathbf{q})$ \hfill\COMMENT{sample splits}
   \STATE $\mathcal{D}_1 \leftarrow \{x_i, y_i | \hat {\mathbf{q}}_i = 1\}$,
          $\mathcal{D}_2 \leftarrow \{x_i, y_i | \hat {\mathbf{q}}_i = 0\}$ \hfill\COMMENT{split data}
   \STATE {\bfseries return} $\mathcal{D}_1, \mathcal{D}_2$  
\end{algorithmic}
\label{algo:eiil}
\end{algorithm}
%%%%%%%%%%%%%%%%%%%%%%%%%%%%%%%%%%%%%%%%%%%%%%%%%%%%%%%%%%%%

Algorithm \ref{algo:eiil} provides pseudocode for the environment inference procedure used in our experiments.

\section{Proofs}\label{sec:proofs}
\subsection{Proof of Proposition \ref{thm:1}}\label{sec:proof_of_thm1}
Consider a dataset with some feature(s) $z$ which are spurious, and other(s) $v$ which are valuable/causal w.r.t. the label $y$.
This includes data generated by models where
$v \rightarrow y \rightarrow z$, such that
$P(y|v,z) = P(y|v)$.
Assume further that the observations $x$ are functions
of both spurious and valuable features: $x := f(v, z)$.
The aim of invariant learning is to form a classifier that predicts $y$ from $x$ that focuses solely on the causal features, i.e., is invariant to $z$ and focuses solely on $v$.

Consider a classifier that produces a score $S(x)$ for example $x$.
In the binary classification setting $S$ is analogous to the model $\Phi$, while the score $S(x)$ is analogous to the representation $\Phi(x)$.
To quantify the degree to which the constraint in the Invariant Principle (\ref{eq:irm}) holds, we introduce a measure called the \emph{group sufficiency gap}\footnote{
This was previously used in a fairness setting by \citet{liu2018implicit} to measure differing calibration curves across groups.
}:
\[\Delta(S,e) = 
\E[
\E[(y|S(x),e_1)] - \E[(y|S(x),e_2)]
]
\]

Now consider the notion of an environment: some setting in which the $x \rightarrow y$ relationship varies (based on spurious features).
Assume a single binary spurious feature $z$.
We restate Proposition 1 as follows:

Claim: If environments are defined based on the agreement of the spurious feature $z$ and the label $y$, then a classifier that predicts based on $z$ alone maximizes the group-sufficiency gap (and vice versa -- if a classifier predicts $y$ directly by predicting $z$, then defining two environments based on agreement of label and spurious feature---$e_1 = \{v,z,y|\indicator{y = z}\}$ and $e_2 = \{v,z,y|\indicator{y \neq z}\}$---maximizes the gap).

We can show this by first noting that if the environment is based on spurious feature-label agreement, then with $e \in \{0, 1\}$ we have $e = \indicator{y=z}$.
If the classifier predicts $z$, i.e. $S(x) = z$, then we have
\[
 \Delta(S,e) 
 = 
 \E[
 \E[y|z(x),\indicator{y=z}] - \E[y|z(x),\indicator{y\neq z}]
 ]
\]

For each instance of $x$ either $z=0$ or $z=1$.
Now we note that when $z=1$ we have ${\E(y|z,\indicator{y=z}) = 1}$ and ${\E(y|z,\indicator{y\neq z}) = 0}$, while when $z=0$  ${E(y|z,\indicator{y=z}]) = 0}$ and ${\E[y|z,\indicator{y \neq z}] = 1}$.
Therefore for each example ${|E(y|z(x),\indicator{y=z}) - E(y|z(x),\indicator{y\neq z}| = 1}$, contributing to an overall $\Delta(S,e)=1$, which is the maximum value for the sufficiency gap.

\subsection{
Heuristic for soft environment assignment based on binning violates the invariance principle
}\label{sec:majmin}
Here we analyze the heuristic discussed in Section \ref{sec:binned-ei}.
We want to show that finding environment assignments in this way both maximizes the violation of the softened version of the regularizer (Equation \ref{eq:ei-obj}), and also also maximally violates the invariance principle (\ref{eq:irm}).

Because the invariance principle $\E[Y|\Phi(X), e] = \E[Y|\Phi(X), e'] \forall e, e'$ is difficult to quantify for continuous $\Phi(X)$, we consider a binned version of the representation, with $b$ denoting the discrete index of the bin in representation space.
Let $q_i \in [0, 1]$ denote the soft assignment of example $i$ to environment 1, and $1 - q_i$ denote its converse, the assignment of example $i$ to environment 2.
Denote by $y_i \in \{0, 1\}$ the binary target for example $i$, and $\hat y \in [0, 1]$ as the model prediction on this example.
Assume that $\ell$ represents a cross entropy or squared error loss so that $\nabla_w \ell(\hat y, y) = (\hat y - y)\Phi(x)$.

Consider the IRMv1 regularizer with soft assignment, expressed as
\begin{align}
    D(q) &= \sum_e ||\nabla_{w|w=1.0} \frac{1}{N_e} \sum_i q_i(e) \ell (w \circ \Phi(x_i), y_i)||^2 \nonumber \\
    &= \sum_e || \frac{1}{N_e} \sum_i q_i(e)(\hat y_i - y_i)\Phi(x_i)||^2 \nonumber \\
    &= ||\frac{1}{\sum_i' q_i'} \sum_{i} q_i(\hat y_i - y_i) \Phi(x_i)||^2 + ||\frac{1}{\sum_i' 1 - q_i'} \sum_{i} (1 - q_i)(\hat y_i - y_i) \Phi(x_i)||^2 \nonumber \\
    &= ||\frac{\sum_i q_i \hat y_i \Phi(x_i)}{\sum_{i'} q_{i'}}
    - \frac{\sum_i q_i y_i \Phi(x_i)}{\sum_{i'} q_{i'}}
    ||^2 + 
    ||\frac{\sum_i (1 - q_i) \hat y_i \Phi(x_i)}{\sum_{i'} 1 - q_{i'}}
    - \frac{\sum_i (1 - q_i) y_i \Phi(x_i)}{\sum_{i'} 1 - q_{i'}}
    ||^2.
\end{align}

Now consider that the space of $\Phi(X)$ is discretized into disjoint bins $b$ over its support, using $z_{i,b} \in \{0, 1\}$ to indicate whether example $i$ falls into bin $b$ according to its mapping $\Phi(x_i)$.
Thus we have 

\begin{align}
    D(q) = &\sum_b (||\frac{\sum_i z_{i,b} q_i \hat y_i \Phi(x_i)}{\sum_{i'} z_{i,b} q_{i'}}
    - \frac{\sum_i z_{i,b} q_i y_i \Phi(x_i)}{\sum_{i'} z_{i,b} q_{i'}}
    ||^2 \nonumber\\
    & + 
    ||\frac{\sum_i z_{i,b} (1 - q_i) \hat y_i \Phi(x_i)}{\sum_{i'} z_{i,b} (1 - q_{i'})}
    - \frac{\sum_i z_{i,b} (1 - q_i) y_i \Phi(x_i)}{\sum_{i'} z_{i,b} (1 - q_{i'})}
    ||^2) 
\end{align}

The important point is that within a bin, all examples have roughly the same $\Phi(x_i)$ value, and the same value for $\hat y_i$ as well.
So denoting $K_b^{(1)} := \frac{\sum_i z_{i,b} q_i \hat y_i \Phi(x_i)}{\sum_{i'} z_{i,b} q_{i'}}$ and $K_b^{(2)} := \frac{\sum_i z_{i,b} (1 - q_i) \hat y_i \Phi(x_i)}{\sum_{i'} z_{i,b} (1 - q_{i'})}$ as the relevant constant within-bin summations, we have the following objective to be maximized by EIIL: 
\[
D(q) = \sum_b( ||K_b^{(1)} -  \frac{\sum_i z_{i,b} q_i y_i \Phi(x_i)}{\sum_{i'} z_{i,b} q_{i'}}||^2 + ||K_b^{(2)} - \frac{\sum_i z_{i,b} (1 - q_i) y_i \Phi(x_i)}{\sum_{i'} z_{i,b} (1 - q_{i'})}||^2.
\]

One way to maximize this is to assign all $y_i=1$ values to environment $1$ ($q_i = 1$ for these examples) and all $y_i=0$ to the other environment $(q_i = 0)$.
We can show this is maximized by considering all of the examples except the $i$-th one have been assigned this way, and then that the loss is maximized by assigning the $i$-th example according to this rule.

Now we want to show that the same assignment maximally violates the invariance principle (showing that this soft EIIL solution provides maximal non-invariance). Intuitively within each bin the difference between $\E[y|e=1]$ and
$\E[y|e=2]$ is maximized (within the bin) if one of these expected label distributions is $1$ while the other is $0$. This can be achieved by assigning all the $y_i=1$ values to the first environment and the $y_i=0$ values to the second.

Thus a global optimum for the relaxed version of EIIL (using the IRMv1 regularizer) also maximally violates the invariance principle.

\subsection{Given CMNIST environments are suboptimal w.r.t. sufficiency gap}\label{sec:suboptimal_sufficiency}
\newcommand{\green}{{\color{ForestGreen} \textsc{Green}}}
\newcommand{\red}{{\color{BrickRed} \textsc{Red}}}
\newcommand{\ez}{e_1}
\newcommand{\eo}{e_2}

The regularizer from IRMv1 encourages a representation for which sufficiency gap is minimized between the available environments.
Therefore when faced with a new task it is natural to measure the natural sufficiency gap between these environments, mediated through a naive or baseline method.
Here we show that for CMNIST, when considering a naive color-based classifier as the reference model, the given environment splits are actually \emph{suboptimal} w.r.t. sufficiency gap, which motivates the inference of environments via EIIL that have a higher sufficiency gap for the reference model.

We begin by computing $ \Delta(S,e) $, the sufficiency gap for color-based classifier $g$ over the given train environments $\{\ez, \eo\}$.
We introduce an auxiliary color variable $z$, which is not observed but can be sampled from via the color based classifier $g$:
\[
p(y|g(x)=x',e) = \E_{p(z|x')} \left[ p(y|z,e,x'). \right]
\]

Denote by $\green$ and $\red$ the set of green and red images, respectively.
I.e. we have $z \in G$ iff $z=1$ and $x \in \green$ iff $z(x) = 1$.
The the sufficiency gap is expressed as
\begin{align}
\Delta(S, e)
&= \E_{p(x,e)} \left[ \Big| \E_{p(y|x,\ez)}[y|g(x), \ez] \nonumber - \E_{p(y|x,\eo)}[y|g(x), \eo] \Big| \right] \nonumber \\
&= \E_{p(z,e)} \left[ \Big| \E_{p(y|z,\ez)}[y|z, \ez] - \E_{p(y|z,\eo)}[y|z, \eo] \Big| \right] \nonumber \\
&= \frac{1}{2}\sum_{z \in \{\green, \red\}} \left[ \Big| \E_{p(y|z,\ez)}[y|z, \ez] - \E_{p(y|z,\eo)}[y|z, \eo] \Big| \right] \nonumber \\
&= \frac{1}{2} (
|\E[y|z=\green,\ez] - \E[y|z=\green,\eo]|
+ |\E[y|z=\red,\ez] - \E[y|z=\red,\eo]|
) \nonumber \\
&= \frac{1}{2}(|0.1 - 0.2| - |0.9 - 0.8|) = \frac{1}{10} \nonumber .
\end{align}

The regularizer in IRMv1 is trying to reduce the sufficiency gap, so in some sense we can think about this gap as a learning signal for the IRM learner.
A natural question would be whether a different set of environment partition $\{e\}$ can be found such that this learning signal is stronger, i.e. the sufficiency gap is increased.
We find the answer is yes. 
Consider an environment distribution $q(e|x ,y,z)$ that assigns each data point to one of two environments.
Any assignment suffices so far as its marginal matches the observed data: $\int_z \int_e q(x,y,z,e) = p^\text{obs}(x,y)$.

We can now express the sufficiency gap (given a color-based classifier $g$) as a function of the environment assignment $q$:
\begin{align}
\Delta(S, e \sim q) &= \E_{q(x,e)} [
|\E_{q(y|x,e,x)}[y|g(x),e_1] - \E_{q(y|x,e,x)}[y|g(x),e_2]|
] \nonumber \\
&= \E_{q(x,e)} [|\E_{q(y|z,e,x)p(z|x)}[y|z,e_1] - \E_{q(y|z,e,x)p(z|x)}[y|z,e_2]| \nonumber
]
\end{align}
Where we use the same change of variables trick as above to replace $g(x)$ with samples from $p(z|x)$ (note that this is the color factor from the generative process $p$ according with our assumption that $g$ matches this distribution).

We want to show that there exists a $q$ yielding a higher sufficiency gap than the given environments.
Consider $q$ that yields the conditional label distribution
\[
q(y|x,e,z) := q(y|e,z) =
\begin{cases}
\indicator{y=z} \ \text{if} \ e = e_1,\\
\indicator{y \neq z} \ \text{if} \ e = e_2.\\
\end{cases}
\]
This can be realized by an encoder/auditor $q(e|x,y,z)$ that ignores image features in $x$ and partitions the example based on whether or not the label $y$ and color $z$ agree.
We also note that $z$ is deterministically the color of the image in the generative process: $p(z|x) = \indicator{x = \red}$

Now we can compute the sufficiency gap:
{\small
\begin{align}
\Delta(S, e \sim q) &= \E_{q(x,e)} [
|\E_{q(y|z,e,x)p(z|x)}[y|z,e_1] - \E_{q(y|z,e,x)p(z|x)}[y|z,e_2]|
] \nonumber \\
&= \frac{1}{2} \E_{x \in \red} |\E_{q(y|z,e,x)p(z|x)}[y|z,e_1] - \E_{q(y|z,e,x)p(z|x)}[y|z,e_2]| \nonumber \\
&\ \ \ \ + \frac{1}{2} \E_{x \in \green} |\E_{q(y|z,e,x)p(z|x)}[y|z,e_1] - \E_{q(y|z,e,x)p(z|x)}[y|z,e_2]| \nonumber \\
&= \frac{1}{2} \E_{x \in \red} (
|\sum_y \sum_z (y * \indicator{y = z} * \indicator{g(x) = z})
- \sum_y \sum_z (y * \indicator{y \neq z} * \indicator{g(x) = z})|
) \nonumber \\
& \ \ \ \  + \E_{x \in \green} \frac{1}{2} (
|\sum_y \sum_z (y * \indicator{y = z} * \indicator{g(x) = z})
- \sum_y \sum_z (y * \indicator{y \neq z} * \indicator{g(x) = z})|
) \nonumber \\
&= \frac{1}{2} \E_{x \in \red} (
|\sum_y (y * \indicator{y = 1} * \indicator{x \in \red})
- \sum_y (y * \indicator{y \neq 1} * \indicator{x \in \red})|
) \nonumber \\
& \ \ \ \  + \E_{x \in \green} \frac{1}{2} (
|\sum_y \sum_z (y * \indicator{y = 0} * \indicator{x \in \green})
- \sum_y \sum_z (y * \indicator{y \neq 0} * \indicator{x \in \green})|
) \nonumber \\
&= \frac{1}{2} \E_{x \in \red}[|1 - 0|] + \E_{x \in \green} [\frac{1}{2} |0 - 1|] = \frac{1}{2} + \frac{1}{2} = 1. \nonumber 
\end{align}
}
Note that $1$ is the maximal sufficiency gap, meaning that the described environment partition maximizes the sufficiency gap w.r.t. the color-based classifier $g$.

\section{Connections Between Invariant Learning and Algorithmic Fairness}\label{sec:fairness_connections}
Here we lay out some connections to algorithmic fairness, where demographic information, which is often considered ``sensitive'', is used to inform learning.
Table \ref{tab:relating_dom_gen_to_fairness} from the main paper provides a high-level comparison of the objectives and assumptions of several relevant methods.
Loosely speaking, recent approaches from both areas share the goal of matching some chosen statistic across a conditioning variable $e$, representing sensitive group membership in algorithmic fairness or an environment/domain indicator in domain generalization.
The statistic in question informs the \emph{learning objective} for the resulting model, and is motivated differently in each case.
In domain generalization, learning is informed by the properties of the test distribution where good generalization should be achieved.
In algorithmic fairness the choice of statistic is motivated by a context-specific \emph{fairness notion}, that likewise encourages a particular solution that achieves ``fair'' outcomes \citep{chouldechova2018frontiers}.

Early approaches to learning fair representations \citep{zemel2013learning,edwards2015censoring,louizos2015variational,zhang2018mitigating,madras2018learning} leveraged statistical independence regularizers from domain adaptation\footnote{
Whereas domain generalization requires model predictions on entirely novel domains at test time, domain adaptation assumes a set of target domain examples are available at test time to guide model adaptation. 
} \citep{ben2010theory,ganin2016domain,tzeng2017adversarial,long2018conditional}, noting that marginal or conditional independence from domain to prediction relates to the fairness notions of demographic parity $\hat y \perp e$ \citep{dwork2012fairness} and equal opportunity $\hat y \perp e | y$ \citep{hardt2016equality}.

Recall that (\ref{eq:irm}) involves an environment-specific conditional label expectation given a data representation ${\E[y|\Phi(x) = h,e]}$.
Objects of this type have been closely studied in the fair machine learning literature, where $e$ now denotes a  ``sensitive'' attribute indicating membership in a protected demographic group (age, race, gender, etc.), and the vector representation $\Phi(x)$ is typically replaced by a scalar score\footnote{
For binary classification, score-based and representation-based approaches are closely related since scores are commonly implemented as (or can be interpreted as) as the linear mapping of a data representation:  $S(x) = w \circ \Phi(x)$.
} $S(x) \in \R$. 
Noting that $\sigma(S(x))$ represents the probability of the model prediction, $\E[y|S(x),e]$ can now be interpreted as a \emph{calibration curve} that must be regulated according to some fairness constraint.
\citet{chouldechova2017fair} showed that equalizing this calibration curve across groups is often incompatible with a common fairness constraint, demographic parity, while \citet{liu2018implicit} studied ``group sufficiency'' of classifiers with strongly convex losses, concluding that ERM naturally finds group sufficient solutions without fairness constraints.

Because \citet{liu2018implicit} consider convex losses, their theoretical results do not hold for neural network representations.
However, by noting the link between group sufficiency and the constraint from (\ref{eq:irm}), we observe that the IRMv1 regularizer (applicable to neural nets) in fact minimizes the group sufficiency gap in the case of a scalar representation $\Phi(x) \subseteq \R$, and when $e$ indicates sensitive group membership.
It is worth noting that \citet{arjovsky2019invariant} briefly discuss using groups as environments, but without specifying a particular fairness criterion.
We leave an empirical study of these methods for future work.

Our approach in searching for worst-case data partitions in EIIL was inspired by recent work on fair prediction without sensitive labels \citep{kearns2018preventing,hebert2017calibration,hashimoto2018fairness,lahoti2020fairness}.
Reliance on sensitive demographic information is cumbersome since it often cannot be collected without legal or ethical repercussions.
\citet{hebert2017calibration} discussed the problem of mitigating subgroup unfairness when group labels are unknown, and proposed \emph{Multicalibration} as a way of ensuring a classifier's calibration curve is invariant to efficiently computable environment splits.
Since the proposed algorithm requires brute force enumeration over all possible environments/groups,
\citet{kim2019multiaccuracy} suggested a more practical algorithm by relaxing the calibration constraint to an accuracy constraint, yielding a \emph{Multiaccurate} classifier.\footnote{
\citet{kearns2018preventing} also proposed a boosting procedure to equalize subgroup errors without sensitive attributes.
}
The goal here is to boost the predictions of a pre-trained classifier through multiple rounds of auditing (searching for worst-case subgroups using an auxiliary model) rather than learning an invariant representation.

A related line of work also leverages inferred subgroup information to improve worst-case model performance using the framework of DRO.
\citet{hashimoto2018fairness} applied DRO to encourage long-term fairness in a dynamical setting where the average loss for a subpopulation influences their propensity to continue engaging with the model.
\citet{lahoti2020fairness} proposed Adversarially Reweighted Learning (ARL), which extends DRO using an auxiliary model to compute the importance weights $\gamma_i$ mentioned above. Amortizing this computation mitigates the tendency of DRO to overfit its reweighting strategy to noisy outliers.

\paragraph{Limitations of generalization-first fairness}
One exciting direction for future work is to apply methods developed in the domain generalization literature to tasks where distribution shift is related to some societal harm that should be mitigated.
However, researchers should be wary of blind ``solutionism'', which can be ineffectual or harmful when the societal context surrounding the machine learning system is ignored \citep{selbst2019fairness}.
Moreover, many aspects of algorithmic discrimination are not simply a matter of achieving few errors on unseen distributions.
Unfairness due to task definition or dataset collection, as discussed in the study of target variable selection by \citet{obermeyer2019dissecting},
may not be reversible by novel algorithmic developments.

\section{Dataset details}\label{sec:dataset_details}
\paragraph{CMNIST}
This dataset was provided by \citet{arjovsky2019invariant}\footnote{\url{https://github.com/facebookresearch/InvariantRiskMinimization}}.
The two training environments comprise $25,000$ images each, with 
$Corr(color, label) = 0.8$ for the firs training environment and $Corr(color, label) = 0.8$ for the second.
A held-out test set with $Corr(color, label)=0.1$ is used for evaluation.
Label noise is applied by flipping the binary target $y$ with probability $\lblnoise = 0.25$, with color correlation applied w.r.t. the noisy label.
Given that only two color channels are used, we follow \citet{arjovsky2019invariant} in downsampling the digit images to $14 \times 14$ pixels and $2$ channels.

\paragraph{Waterbirds}
We follow the procedure outlined by \citet{sagawa2019distributionally} to reproduce the Waterbirds dataset. 
As noted by the authors, due to random seed differences our version of the dataset may differ slightly from the one originally used by the paper.
The train/validation/test splits are of size $4,795$/$1,200$/$5,794$.
As noted in the Appendix of \citep{sagawa2019distributionally}, the validation and test distributions represent upweight the minority groups so that the number of examples coming from each habitat is equal (although there are still marginally more landbirds than waterbirds). For example on train set the subgroup sizes are $3,498$/$184$/$56$/$1,057$ while on the test set the sizes are $467$/$466$/$133$/$133$.

\paragraph{CivilComments-WILDS}
We use the train/validation/test splits from \citet{koh2020wilds}; we refer the interested reader the Appendix of their paper for a detailed description of this version of the dataset, including how it differs from the original dataset \citep{borkan2019nuanced}.

\paragraph{Constructing the Adult-Confounded dataset}
To create our semi-synthetic dataset, called Adult-Confounded, we start by observing that the conditional distribution over labels varies across the subgroups, and in some cases subgroup membership is very predictive of the target label.
We construct a test set (a.k.a. the audit set) where this relationship between subgroups and target label is reversed.

The four sensitive subgroups are defined following the procedure of \citet{lahoti2020fairness}, with sex (recorded as binary: Male/Female) and binarized race (Black/non-Black) attributes compose to make four possible subgroups: Non-Black Males (SG1), Non-Black Females (S2), Black Males (SG3), and Black Females (SG4).

We start with the observation that each subgroup has a different correlation strength with the target label, and in some cases subgroup membership alone can be used to achieve relatively low error rates in prediction.
As these correlations should be considered ``spurious'' to mitigate unequal treatment across groups, we create a semi-synthetic variant of the UCI Adult dataset, which we call Adult-Confounded, where these spurious correlations are exaggerated. 
Table \ref{tab:confounded_adult_stats} shows various conditional label distributions for the original dataset and our proposed variant.
The test set for Adult-Confounded reverses the correlation strengths, which can be thought of as a worst-case audit to ensure the model is not relying on subgroup membership alone in its predictions.
We generate samples for Adult-Confounded using importance sampling, keeping the original train/test splits from UCI Adult as well as the subgroup sizes, but sampling individual examples under/over-sampled according to importance weights $\frac{p^{Adult-Confounded}}{p^{UCIAdult}}$.

\begin{table}[ht]
\centering
\begin{tabular}{*5c}
\toprule
Subgroup ($SG$) &  \multicolumn{4}{c}{$p(y=1|SG)$}\\
\midrule
&  \multicolumn{2}{c}{UCIAdult} & \multicolumn{2}{c}{Adult-Confounded}\\
{}   & Train   & Test    & Train   & Test\\
1   &  0.31 & 0.30   & 0.94  & 0.06\\
2   &  0.11 & 0.12   & 0.06  & 0.94\\
3   &  0.19  & 0.16   & 0.94  & 0.06\\
4   &  0.06  & 0.04   & 0.06  & 0.94\\
\bottomrule
\end{tabular}
\caption{
Adult-Confounded is a variant of the UCI Adult dataset that emphasizes test-time distribution shift.
}
\label{tab:confounded_adult_stats}
\end{table}

\section{Experimental details}\label{sec:experimental_details}
\paragraph{Model selection}
\citet{krueger2020out} discussed the pitfalls of achieving good test performance on CMNIST by using test data to tune hyperparameters.
Because our primary interest is in the properties of the inferred environment rather than the final test performance, we sidestep this issue in the Synthetic Regression and CMNIST experiments by using the default parameters of IRM without further tuning.
However for Adult-Confounded a specific strategy for model selection is needed.

We refer the interested reader to \citet{gulrajani2020search} for an extensive discussion of possible model selection strategies. They also provide a large empirical study showing that ERM is difficult baseline to beat when all methods are put on equal footing w.r.t. model selection.

In our case, we use the most relaxed model selection method proposed by \citet{gulrajani2020search}, which amounts to allowing each method a $20$ test evaluations using hyperparameter chosen at random from a reasonable range, with the best hyperparameter setting selected for each method.
While none of the methods is given an unfair advantage in the search over hyperparameters, the basic model selection premise does not translate to real-world applications, since information about the test-time distribution is required to select hyperparameters.
Thus these results can be understood as being overly optimistic for each method, although the relative ordering between the methods can still be compared.

\paragraph{Training times}

Because EIIL requires a pre-trained reference model and optimization of the EI objective, overall training time is longer than standard invariant learning.
It depends primarily on the number of steps used to train the reference model and number of steps used in EI optimization. 
The extra training time incurred is manageable and varies from dataset to dataset.

In CMNIST, we train the ERM reference model for $1,000$ steps, which is the same duration as the downstream invariant learner that eventually uses the inferred environments.
In this setting the $10,000$ steps required to optimize the EI objective is actually more than used for representation learning.
The overall EIIL train time is $6.6$ minutes to run 10 restarts on a NVIDIA Tesla P100, compared with $2.18$ minutes for ERM and $2.20$ minutes for IRM.

However, as the problem size scales, the relative overhead cost of EIIL becomes progressively discounted.
On Waterbirds, training GroupDRO takes $4.716$ hours on a NVIDIA Tesla P100.
Our reference model trains for $1$ epoch, so taking this into account along with the $20,000$ steps of EI optimization, EIIL runs at 4.737 hours.
This is a relative increase of 0.4\%.

\paragraph{Batch environment inference}
As mentioned in the main paper, we aggregate logits for the entire training set and optimize the EI objective using the entire training batches.
This can be done by cycling through the train set once in minibatches, computing logits per minibatch, and aggregating the logits only (discarding network activations) prior to EI.
We leave minibatched environment inference and amortization of the soft environment assignments to future work.

\paragraph{Experimental infrastructure}
Our experiments were run on a cluster of NVIDIA Tesla P100 machines.

\paragraph{CMNIST}
IRM is trained on the two training environments and tested on a holdout environment constructed from $10,000$ test images in the same way as the training environments, where colour is predictive of the noisy label 10\% of the time. So using color as a feature to predict the label will lead to an accuracy of roughly 10\% on the test environment, while it yields 80\% and 90\% accuracy respectively on the training environments. 

To evaluate EIIL we remove the environment identifier from the training set and thus have one training set comprised of $50,000$ images from both original training environments. We then train an MLP with binary cross-entropy loss on the training environments, freeze its weights and use the obtained model to learn environment splits that maximally violate the IRM penalty.
When optimizing the inner loop of EIIL, we use Adam with learning rate 0.001 for $10,000$ steps with full data batches used to computed gradients.

The obtained environment partitions are then used to train a new model from scratch with IRM.
Following \citet{arjovsky2019invariant}, we allow the representation to train for several hundred annealing steps before applying the IRMv1 penalty.

We used the default architecture---an MLP with two hidden layers of $390$ neurons---and hyperparameter values\footnote{\url{https://github.com/facebookresearch/InvariantRiskMinimization/blob/master/code/colored_mnist/reproduce_paper_results.sh}}----learning rate, weight decay, and penalty strength---from \citep{arjovsky2019invariant}.
We do not use minibatches as the entire dataset fits into memory.

\paragraph{Waterbirds}
Following \citet{sagawa2019distributionally}, we use the default \texttt{torchvision} ResNet50 models, using the pre-trained weights as the initial model parameters, and train without any data augmentation using the 
For GroupDRO and ERM, we use hyperparameters reported by the authors\footnote{\url{https://worksheets.codalab.org/worksheets/0x621811fe446b49bb818293bae2ef88c0}}, and note that the authors make use of the the validation set (whose distribution contains less group imbalance than the training data), to select hyperparameters in their experiments (all methods benefit equally from this strategy).
We train for $300$ epochs without any early stopping (to avoid any further influence from the validation data).
For EIIL, we optimize the EI objective of EIIL with learning rate $0.01$ for $20,000$ steps using the Adam optimizer, and use GroupDRO (using the same hyperparameters as the GroupDRO baseline) as the invariant learner.
An ERM model trained for $1$ epoch was used as the reference model.
We also tried using reference modeled trained for longer, but found that EIIL did not perform as well in this case.
We hypothesize that this is because the reference ERM model focuses on background features early in training, leading to stark performance discrepancies across subgroups, which in turn provides a strong learning signal for EIIL to infer effective environments.
While subgroup disparities are present for more well-trained models, the learning signal in the EI phase will weaken.

\paragraph{Adult-Confounded}
Following \citet{lahoti2020fairness}, we use a two-hidden-layer MLP architecture for all methods, with $64$ and $32$ hidden units respectively, and a linear adversary for ARL.
We use IRM as the invariant learner in the final stage of EIIL.
We optimize all methods using Adagrad; learning rates, number of steps, and batch sizes chosen by the model selection strategy described above (with $20$ test evaluations per method), as are penalty weights for IRMv1 regularizer and standard weight decay.
For the inner loop of EIIL (inferring the environments), we use the same settings as in CMNIST.
We find that the performance of EIIL is somewhat sensitive to the number of steps taken with the IRMv1 penalty applied.
To limit the number of test queries needed during model selection, we use an early stopping heuristic by enforcing the IRMv1 penalty only during the final 500 steps of training, with the previous steps serving as annealing period to learn a baseline representation to be regularized.

\paragraph{CivilComments-WILDS}
Following \citep{koh2020wilds}, we finetune DistilBERT embeddings \citep{sanh2019distilbert} using the default HuggingFace implementation and default weights \citep{wolf2019huggingface}.
EIIL uses an ERM reference classifier and its inferred environments are fed to a GroupDRO invariant learner.
During prototyping the EI step, we noticed that the binning heuristic described in Section \ref{sec:binned-ei} consistently split the training examples into environments according to the error cases of the reference classifier.
Because error splitting is even simpler to implement than confidence binning, we used this heuristic for the EI step; we believe this is a promising approach for scaling EI to large datasets, and note its equivalence to the first stage of the method independently proposed by \citet{liu2021just}, which is published concurrently to ours.
We experimented with gradient-based EI on this dataset, but did not find any improvement over the (faster) heuristic EI.

On this dataset, we treat reference model selection as part of the overall model selection process, meaning that the hyperparameters of the ERM reference model are treated as a subset of the overall hyperparameters tuned during model selection.
Specifically we used a grid search to tune the reference model learning rate (1e-5, 1e-4), optimizer type (Adam, SGD) and scheduler (linear, plateau), and gradient norm clamping (off, clamped at 1.0), as well as the invariant learner (GroupDRO) learning rate (1e-5, 1e-4).
Moreover, we allow all methods to evaluate worst-group validation accuracy to tune these hyperparameters; such validation data will not be available in most settings, so this result can be seen as an optimistic view of the performance of all methods, including EIIL.
We train all methods (including the reference model) for $5$ epochs, with the best epoch chosen according to validation performance.
Interestingly, the reference model chosen in this way was a constant classifier, so the overall EIIL solution is equivalent to GroupDRO using the class label as the environment label.

The oracle GroupDRO method trains on two environments, with one containing comments where \emph{any} of the $8$ sensitive groups was mentioned, and other environment containing the remaining comments.
We experimented with allowing the oracle method access to more fine-grained environment labels by evaluating all $2^8$ combinations of binary group labels, but did not find any significant performance boost (consistent with observations from \citet{koh2020wilds}).

\section{Additional Empirical Results}\label{sec:additional_results}

\subsection{Synthetic Data}\label{sec:synthetic-regression}

\begin{table}[ht]
\centering
\begin{tabular}{lll}
\toprule
{} &        Causal MSE &     Noncausal MSE \\
\midrule
\textsc{ERM}                     &  0.827 $\pm$ 0.185 &  0.824 $\pm$ 0.013 \\
ICP &  1.000 $\pm$ 0.000 &  0.756 $\pm$ 0.378 \\
IRM &  0.666 $\pm$ 0.073 &  0.644 $\pm$ 0.061 \\
\textbf{EIIL}     &  \textbf{0.148} $\pm$ \textbf{0.185} &  \textbf{0.145} $\pm$ \textbf{0.177} \\
\bottomrule
\end{tabular}
\caption{IRM using EIIL-discovered environments ($e_{\textsc{EIIL}}$) outperforms IRM in a synthetic regression setting without the need for hand-crafted environments ($e_{\textsc{HC}}$). This is because the reference representation $\tilde \Phi = \Phi_{\textsc{ERM}}$ uses the spurious feature for prediction.
MSE +\- standard deviation across 5 runs reported.}
\label{tab:sem}
\end{table}

We begin with a regression setting originally used as a toy dataset for evaluating IRM \citep{arjovsky2019invariant}.
The features $\obs \in \spa$ comprise a ``causal'' feature $\cf \in \hsp$ concatenated with a ``non-causal'' feature $\ncf \in \hsp$: $\obs = [\cf, \ncf]$.
Noise varies across hand-crafted environments $e$:
\begin{align*}
     \cf &= \epsilon_\cf & \epsilon_\cf \sim \Normal(0, 25) \\
     \lbl &= \cf + \epsilon_\lbl & \epsilon_\lbl \sim \Normal(0, e^2) \\
     \ncf &= \lbl + \epsilon_\ncf & \epsilon_\ncf \sim \Normal(0, 1).
\end{align*}

We evaluated the performance of the following methods:
\begin{itemize}
    \item \textbf{ERM:} A naive regressor that does not make use of environment labels $e$, but instead optimizes the average loss on the aggregated environments;
    \item \textbf{IRM:} the method of \citet{arjovsky2019invariant} using hand-crafted environment labels;
    \item \textbf{ICP:} the method of \citet{peters2016causal} using hand-crafted environment labels;
    \item \textbf{EIIL:} our proposed method (which does use hand-crafted  environment labels) that infers useful environments based on the naive ERM, then applies IRM to the inferred environments.
\end{itemize}
    
The regression methods fit a scalar target $y = \mathbf{1}^T \lbl$ via a regression model $\hat y \approx \mathbf{w}^T \obs$ to minimize $||y - \hat y||$ w.r.t. $\mathbf{w}$, plus an invariance penalty as needed.
The optimal (causally correct) solution is $\mathbf{w}^* = [\mathbf{1}, \mathbf{0}]$
Given a solution $[\pcf, \pncf]$ from one of the methods, we report the mean squared error for the causal and non-causal dimensions as $||\pcf - \mathbf{1}||_2^2$ and $||\pncf - \mathbf{0}||_2^2$ (Table \ref{tab:sem}).
Because $\cf$ is marginally noisier than $\ncf$, ERM \ focuses on the spurious $\ncf$.
IRM using hand-crafted environments, denoted IRM,
exploits variability in noise level in the non-causal feature (which depends on the variability of $\sigma_\lbl$) to achieve lower error.
Using EIIL instead of hand crafted environments yields an improvement on the resulting IRM solution by learning worst-case environments for invariant training.

\begin{figure*}[ht]
\centering
\subfigure[Causal error]{
\includegraphics[width=.45\textwidth]{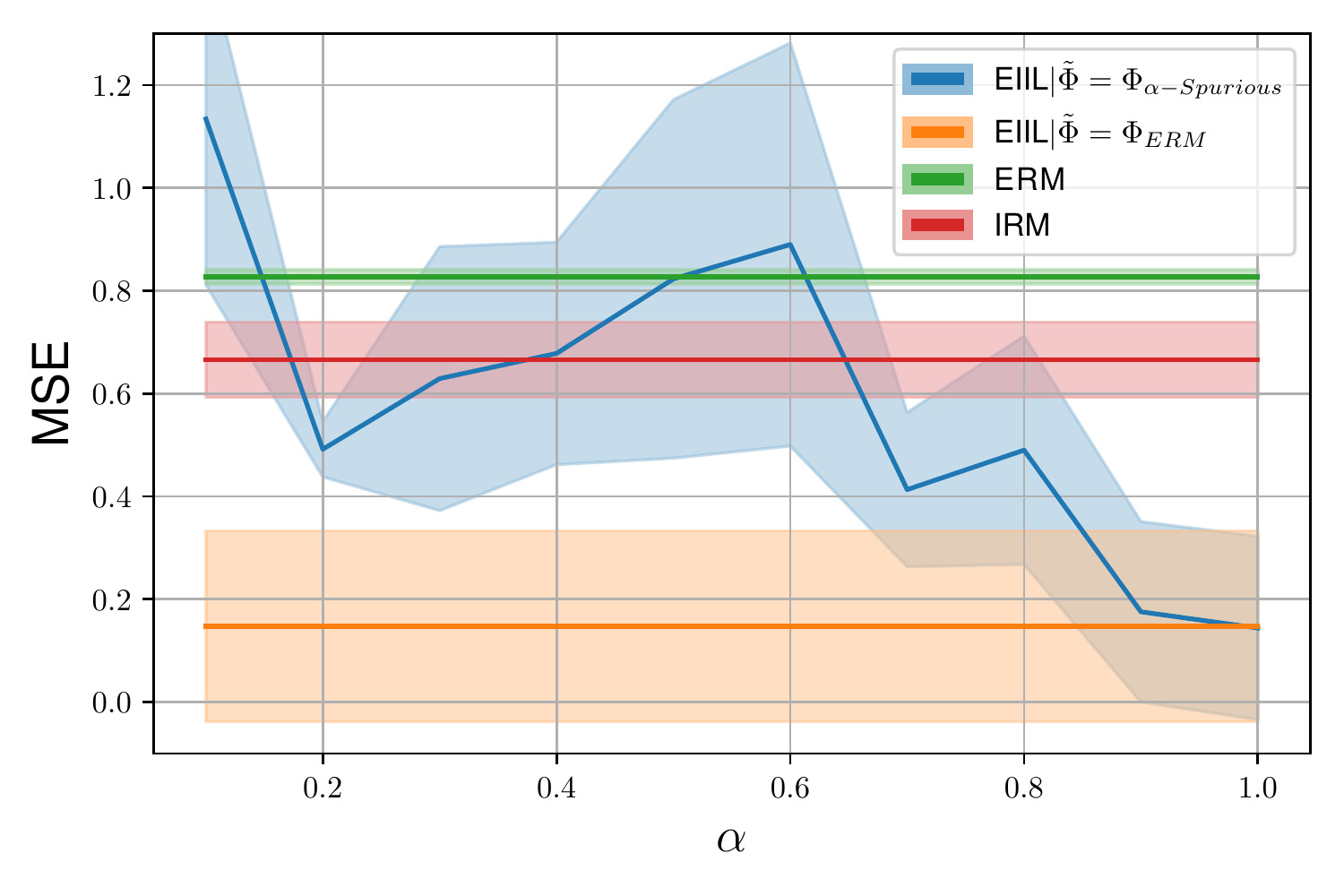}
}
\subfigure[Non-causal error]{
\includegraphics[width=.45\textwidth]{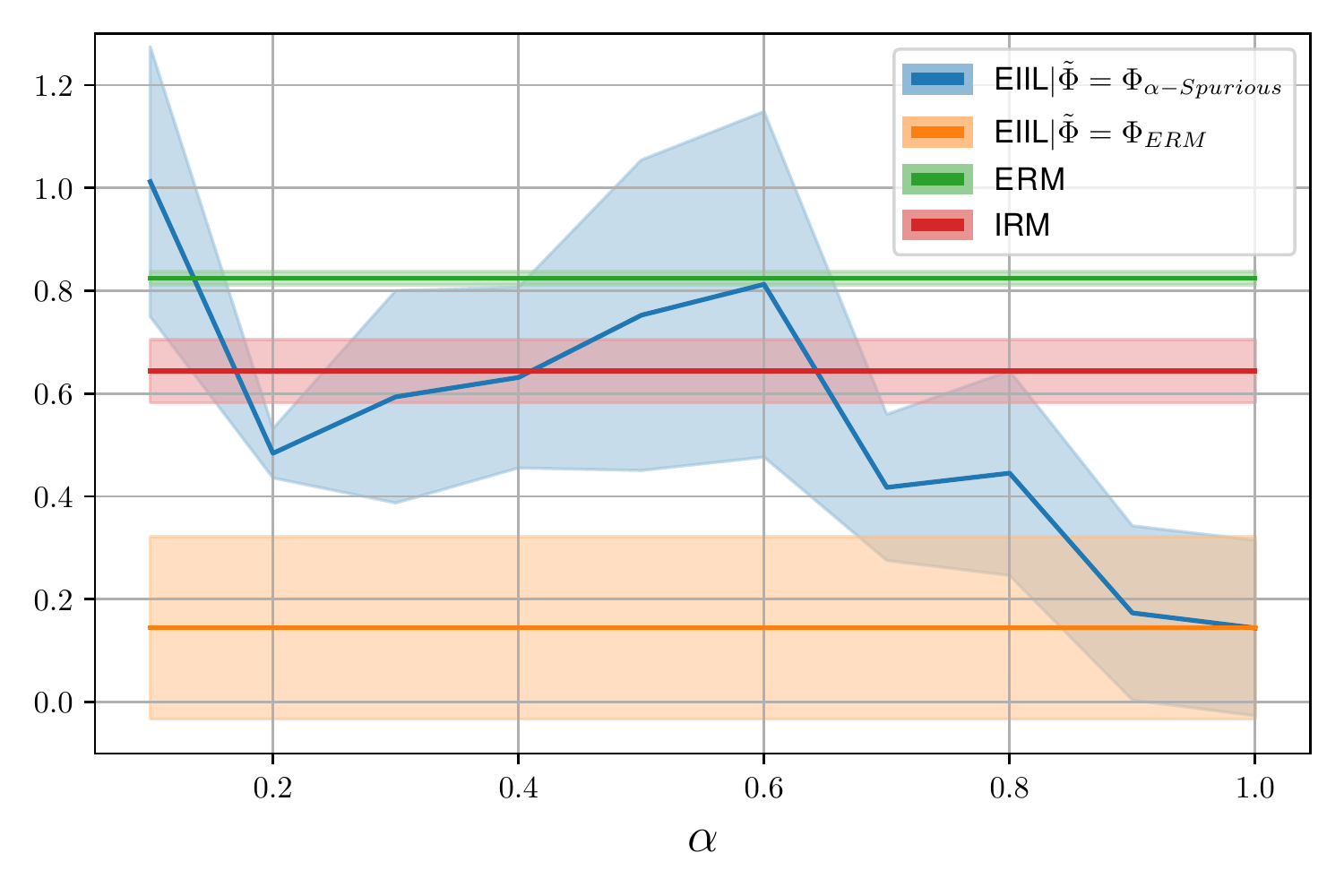}
}
\label{fig:sem_causal_err}
\caption[]{MSE of the causal feature $\cf$ and non-causal feature $\ncf$. EIIL applied to the ERM solution (Black) 
out-performs IRM based on the hand-crafted environment (Green vs. Blue).
To examine the inductive bias of the reference model $\tilde \Phi$, we hard code a model $\tilde \Phi_{\alpha-\textsc{Spurious}}$ where
$\alpha$ controls the degree of spurious feature representation in the reference classifier;
EIIL outperforms IRM when the reference $\tilde \Phi$ focuses on the spurious feature, e.g. with $\tilde \Phi$ as ERM or $\alpha$-\textsc{Spurious} for high $\alpha$.
}
\label{fig:sem_alpha_sweep}
\end{figure*}

We show in a follow-up experiment that the EIIL solution is indeed sensitive to the choice of reference representation, and in fact, can only discover useful environments (environments that allow EIIL to learn the correct causal representation) when the reference representation encodes the \emph{incorrect} inductive bias by focusing on the spurious feature.
We can explore this dependence of EIIL on the mix of spurious and non-spurious features in the reference model by constructing a $\tilde \Phi$ that varies in the degree it focuses on the spurious feature, according to convex mixing parameter $\alpha \in [0, 1]$.
${\alpha=0}$ indicates focusing entirely on the correct causal feature, while ${\alpha=1}$ indicates focusing on the spurious feature.
We refer to  this variant as 
EIIL$|\tilde \Phi = \Phi_{\alpha-\textsc{Spurious}}$,
and measure its performance as a function of $\alpha$ (Figure \ref{fig:sem_alpha_sweep}).
Environment inference only yields good test-time performance for high values of $\alpha$, where the reference model captures the \emph{incorrect} inductive bias.

\subsection{ColorMNIST}\label{sec:cmnist_extra_results}

\begin{table}[ht]
\centering
\begin{tabular}{lll}
\toprule
{}                                                  &     Train accs &      Test accs \\
\midrule
Grayscale (oracle)                                  & 75.3 $\pm$ 0.1 & 72.6 $\pm$ 0.6 \\
IRM (oracle envs)                                   & 71.1 $\pm$ 0.8 & 65.5 $\pm$ 2.3 \\
\midrule
ERM                                                 & 86.3 $\pm$ 0.1 & 13.8 $\pm$ 0.6 \\
EIIL                                                & 73.7 $\pm$ 0.5 & 68.4 $\pm$ 2.7 \\
Binned EI heuristic (Sec. \ref{sec:binned-ei})      & 73.9 $\pm$ 0.5 & 69.0 $\pm$ 1.5 \\
$\Phi_{Color}$                                      & 85.0 $\pm$ 0.1 & 10.1 $\pm$ 0.2 \\
EIIL$|\tilde \Phi = \Phi_{Color}$                   & 75.9 $\pm$ 0.4 & 68.0 $\pm$ 1.2 \\
ARL                                                 & 88.9 $\pm$ 0.2 & 20.7 $\pm$ 0.9 \\
GEORGE& 84.6 $\pm$ 0.3 & 12.8 $\pm$ 2.0 \\
LFF;$\mathcal{L}_{bias}=\text{GCE}_{q\rightarrow0}$ & 96.6 $\pm$ 1.3 & 30.6 $\pm$ 1.0\\
LFF;$\mathcal{L}_{bias}=\text{GCE}_{q=0.7}$         & 15.0 $\pm$ 0.1 & 90.0 $\pm$ 0.3\\
\bottomrule
\end{tabular}
\caption{
Additional baselines for the CMNIST experiment reported in Table \ref{tab:table_teaser}.
The mean and standard deviation of accuracy across ten runs ($\lblnoise=0.25$) are reported.
See text for description of the baseline methods.
}
\label{tab:cmnist_acc}
\end{table}
Table \ref{tab:cmnist_acc} expands on the results from Table \ref{tab:table_teaser} by adding the following baselines that do not require environment labels:
\begin{itemize}
    \item Grayscale: a classifier that removes color via pre-processing, which represents an oracle solution
    \item EIIL$|\tilde \Phi = \Phi_{ERM}$ (reported as EIIL in Table \ref{tab:table_teaser}
    \item Binned EI heuristic: the binning heuristic for environment inference described in Section \ref{sec:binned-ei}.
    \item $\Phi_{Color}$: a hard-coded classifier that predicts \emph{only} based on the digit color
    \item EIIL$|\tilde \Phi = \Phi_{Color}$: EIIL using color-based classifier (rather than $\Phi_{ERM}$) as reference.
    \item GEORGE \citep{sohoni2020no}: This two-stage method seeks to learn the ``hidden subclasses'' by fitting a latent cluster model to the (per-class) distribution of logits of a reference model. The inferred hidden subclasses are fed to a GroupDRO learner, so this approach can be seen as an instance of EIIL under particular choices of (unsupervised) EI and (robust optimization) IL objectives.
    \item ARL \citep{lahoti2020fairness}: A variant of DRO that uses an adversary/auxiliary model to learn worst-case per-example importance weights. Unlike with EIIL, the auxiliary model and main model are trained jointly.
    \item LFF \citep{nam2020learning} jointly trains a ``biased'' model $f_B$ and ``debiased'' model $f_D$. $f_B$ is similar to our ERM reference model, but is trained with $\text{GCE}_q(p(x;\theta), y) = \frac{1-p_y(x;\theta)^q}{q}$ with hyperparameter $q \in (0, 1]$,\footnote{as $q \rightarrow 0$ GCE becomes standard cross entropy}
and its per-example losses determine importance weights for $f_D$.
\end{itemize}

When expanding this study we find that, unlike EIIL, the new baselines fail to find an invariant classifier that predicts based on shape rather than color.
Given that GEORGE does a type of unsupervised EI, it is perhaps surprising that it cannot uncover optimal environments for use with its GroupDRO learner. 
We hypothesize that this is due to assumption of the relevant latent environment labels being ``hidden subclasses'', meaning that all examples in an optimal environment must share the same class label value.
In the CMNIST dataset, this assumption does not hold due to label noise.

We find that, on this dataset, LFF is very sensitive to the hyperparameter $q$, which shapes the GCE loss of $f_B$.
Interestingly, using the default value of $q=0.7$, LFF performs optimally on the test set, but this is \emph{not} because the method has learned an invariant classifier based on the digit shape.
The below-chance train set performance reveals that LFF has learned an \emph{anti-color} classifier, exactly the opposite of what ERM does.
When $q$ approaches zero (GCE approaches standard cross entropy), LFF fails to generalizes to the OOD test distribution.

Finally, we found that because the reference classifier predicts with high confidence on the training set, there are only two populated bins in practice.
Consequentially, the binned EI heuristic is equivalent to splitting errors into one environment and correct predictions into the other.

\subsection{Adult-Confounded}\label{sec:ablation}
\paragraph{Subgroup sufficiency}
In the main result we showed that EIIL improves test calibration and accuracy our variant of the UCIAdult dataset.
Because the test set is subject to a drastic distribution shift where the correlation pattern between subgroup membership and label is reversed relative to the training set, we can say that this robustness in performance suggests that EIIL does not rely on subgroup membership to make its predictions.

%%%%%%%%%%%%%%%%%%%%%%%%%%%%%%%%%%%%%%%%%%%%%%%%%%%%%%
% UCIAdult-Confounded extra calibration plots
%%%%%%%%%%%%%%%%%%%%%%%%%%%%%%%%%%%%%%%%%%%%%%%%%%%%%%
\begin{figure}[ht]
\newcommand{\mywidth}{.3\textwidth}
\centering
\subfigure[]{\includegraphics[width=\mywidth]{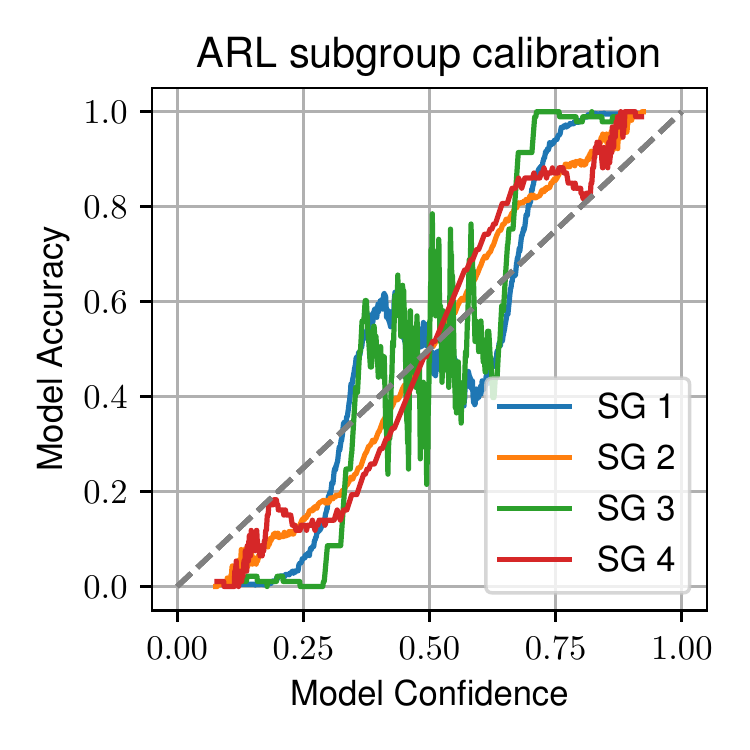}}
\subfigure[]{\includegraphics[width=\mywidth]{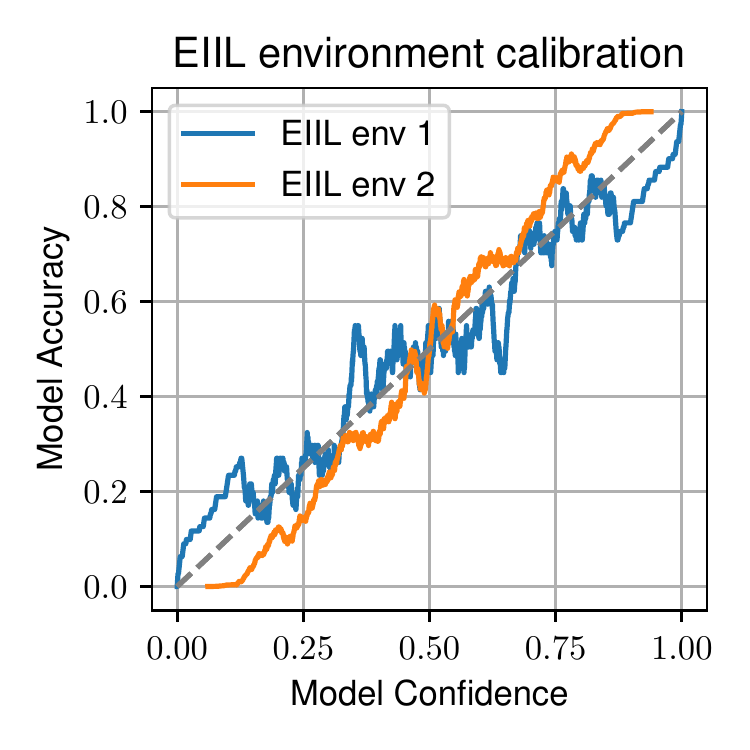}}
\subfigure[]{\includegraphics[width=\mywidth]{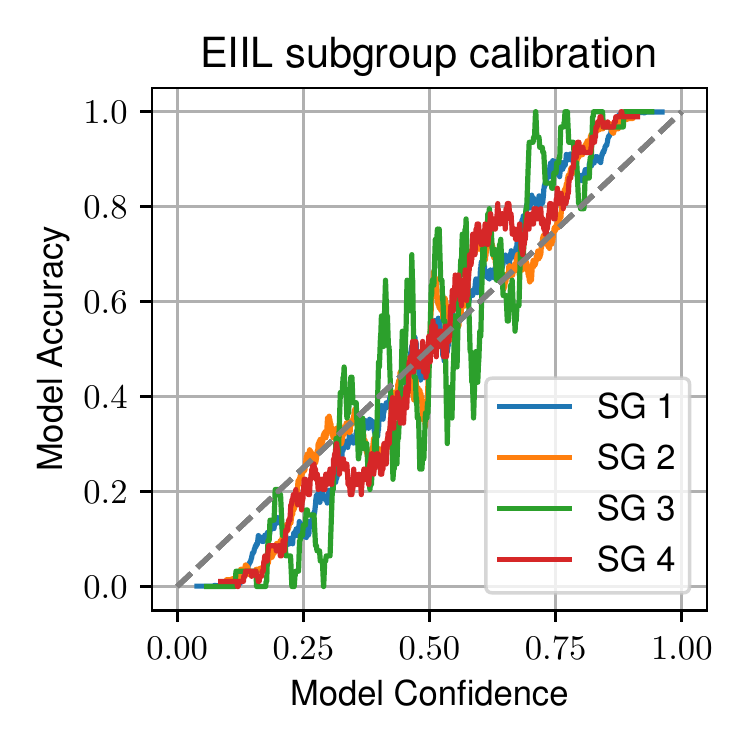}}

\caption{
We examine \emph{subgroup sufficiency}---whether calibration curves match across demographic subgroups---on the Adult-Confounded dataset.
Whereas ARL is not subgroup-sufficient (a), EIIL infers worst-case environments and regularizes their calibration to be similar (b), ultimately improving subgroup sufficiency (c).
Note that neither method uses sensitive group information during learning.
}
\label{fig:group_sufficiency_extra}
\end{figure}
%%%%%%%%%%%%%%%%%%%%%%%%%%%%%%%%%%%%%%%%%%%%%%%%%%%%%%

Beyond the global calibration profile, we can also examine calibration curves for the various subgroups, noting again that subgroup labels were not used to train EIIL or the ARL baseline.
Figure \ref{fig:group_sufficiency_extra} shows the calibration profiles on the training data.
We find that ARL contains noticeable discrepancies in the calibration curves across groups indicating that subgroup sufficiency has not been achieved.
EIIL infers environments during the EI phase, which are then implicitly regularized to have roughly the same calibration profile during invariant learning.
This can be seen by examining the calibration plots for the training data when it is stratified into the two inferred environments.
Finally, looking at calibration curves for the subgroups themselves suggests that EIIL has improved on subgroup sufficiency relative to ARL by better matching the calibration curves across subgroups.
These curves still exhibit some noise, indicating that further progress on subgroup sufficiency could be made by changing the invariant learner, possibly by using a different regularize (besides IRMv1) that better enforces the invariance principle.

\paragraph{Ablation}
Here we provide an ablation study extending Adult-Confounded experiments to demonstrate that both ingredients in the EIIL solution---finding worst-case environment splits and regularizing using the IRMv1 penalty---are necessary to achieve good test-time performance on the Adult-Confounded dataset.

\begin{table}[ht]
\centering
\begin{tabular}{lll}
\toprule
{} &      Train accs &       Test accs \\
\midrule
EIIL                  &  68.7 $\pm$ 1.7 &  \textbf{79.8 $\pm$ 1.1} \\
EIIL (no regularizer) &  78.6 $\pm$ 2.0 &  69.2 $\pm$ 2.8 \\
IRM (random environments)     &  \textbf{94.7 $\pm$ 0.1} &  17.6 $\pm$ 1.6 \\
\bottomrule
\end{tabular}
\caption{
Our ablation study shows that both ingredients of EIIL (finding worst-case environments and regularizing invariance across them) are required to achieve good test-time performance on the Adult-Confounded dataset.
}
\label{tab:ablation}
\end{table}

From \citet{lahoti2020fairness} we see that ARL can perform favorably compared with DRO \citep{hashimoto2018fairness} in adaptively computing how much each example should contribute to the overall loss, i.e. computing the per-example $\gamma_i$ in $C = \E_{x_i, y_i \sim p} [ \gamma_i \ell(\Phi(x_i), y_i) ]$.
Because all per-environment risks in IRM are weighted equally (see Equation \ref{eq:irmv1}), and each per-environment risk comprises an average across per-example losses within the environment, each example contributes its loss to the overall objective in accordance with the size of its assigned environment.
For example with two environments $e_1$ and $e_2$ of sizes $|e_1|$ and $|e_2|$, we implicitly have the per-example weights of $\gamma_i = \frac{1}{|e_1|}$ for $i \in e_1$ and $\gamma_i = \frac{1}{|e_2|}$ for $i \in e_2$, indicating that examples in the smaller environment count more towards the overall objective.
Because EIIL can discover worst-case environments of unequal sizes, we measure the performance of EIIL using only this reweighting, without adding the gradient-norm penalty typically used in IRM (i.e. setting $\lambda=0$).
To determine the benefit of worst-case environment discovery, we also measure IRM with random assignment of environments.
Table \ref{tab:ablation} confirms that both ingredients are required to attain good performance using EIIL.

\end{document}

%% file: math_commands.tex
\usepackage{amsmath,amsfonts,bm}

\newcommand{\indicator}[1]{\mathds{1} (#1)}  % requires dsfont

\def\eqref#1{equation~\ref{#1}}
\def\1{\bm{1}}

\DeclareMathAlphabet{\mathsfit}{\encodingdefault}{\sfdefault}{m}{sl}
\SetMathAlphabet{\mathsfit}{bold}{\encodingdefault}{\sfdefault}{bx}{n}

\newcommand{\E}{\mathbb{E}}

\newcommand{\R}{\mathbb{R}}

\DeclareMathOperator*{\argmax}{arg\,max}
\DeclareMathOperator*{\argmin}{arg\,min}